\documentclass[11pt,a4paper]{article}
\usepackage[hyperref]{emnlp2020}
\usepackage{times}
\usepackage{multirow}
\usepackage{latexsym}
\usepackage{amsmath}
\usepackage{booktabs}
\usepackage{graphicx}
\usepackage{pifont}
\usepackage{array}
\usepackage{enumitem}
\usepackage{subcaption}
\usepackage{bm}

\usepackage[ruled,vlined]{algorithm2e}

\usepackage{soul}
\usepackage{afterpage}

\usepackage{microtype}

\aclfinalcopy %
\DeclareMathOperator*{\argmax}{arg\,max}

\newcommand{\cmark}{\ding{51}}%
\newcommand{\xmark}{\ding{55}}%

\newif\ifhidecomments
\hidecommentsfalse

\ifhidecomments
    \newcommand{\sam}[1]{}
    \newcommand{\chenhao}[1]{}
    \newcommand{\ani}[1]{}
\else
    \newcommand{\chenhao}[1]{\textcolor{cyan}{\textsc{\textbf{[#1 --ct]}}}}
    \newcommand{\sam}[1]{\textcolor{magenta}{\textsc{\textbf{[#1 --sc]}}}}
    \newcommand{\ani}[1]{\textcolor{blue}{\textsc{\textbf{[#1 --ani]}}}}
\fi

\newcommand{\para}[1]{\noindent{\bf #1}}
\newcommand{\figref}[1]{Fig.~\ref{#1}}

\newcommand{\secref}[1]{\S\ref{#1}}
\newcommand{\tableref}[1]{Table~\ref{#1}}
\newcommand{\roberta}{RoBERTa\xspace}
\newcommand{\multirc}{MultiRC\xspace}
\newcommand{\wikiattack}{WikiAttack\xspace}
\newcommand{\esnli}{E-SNLI\xspace}
\newcommand{\sst}{SST\xspace}
\newcommand{\movie}{Movie\xspace}
\newcommand{\movies}{Movie\xspace}
\newcommand{\nontoxic}{no-attack\xspace}
\newcommand{\toxic}{personal-attack\xspace}
\newcommand{\fever}{FEVER\xspace}
\newcommand{\qastyle}{document/query-style\xspace}
\newcommand{\rationale}{{\bm \alpha}}
\newcommand{\data}{{\bm x}}

\newcommand{\highlightrationale}[1]{\underline{\bf{#1}}\xspace}
\newcommand{\problematic}[1]{\textcolor{red}{#1}\xspace}

\newcommand{\norationale}{\textit{No rationale}\xspace}
\newcommand{\evalrationale}{\textit{Eval rationale}\xspace}
\newcommand{\trainevalrationale}{\textit{Train-eval rationale}\xspace}

\title{Evaluating and Characterizing Human Rationales}

\author{
    Samuel Carton\thanks{\xspace\xspace Equal contribution.}, Anirudh Rathore\footnotemark[1], Chenhao Tan\\ 
    University of Colorado Boulder \\
   \texttt{\{samuel.carton,anirudh.rathore,chenhao.tan\}@colorado.edu}
}

\date{}

\begin{document}
\maketitle
\begin{abstract}

Two main approaches for evaluating the quality of machine-generated rationales are: 1) using human rationales as a gold standard; and 2) automated metrics based on how rationales affect model behavior. An open question, however, is how human rationales fare with these automatic metrics. Analyzing a variety of datasets and models, we find that human rationales do not necessarily perform well on these metrics. To unpack this finding, we propose improved metrics to account for model-dependent baseline performance. We then propose two methods to further characterize rationale quality, one based on model retraining and one on using ``fidelity curves'' to reveal properties such as irrelevance and redundancy. Our work leads to actionable suggestions for evaluating and characterizing rationales.
\end{abstract}

\section{Introduction}

Explanations in NLP
often take the form of \textit{rationales}, subsets of input tokens that are considered important to the model's decision \citep{deyoung2019eraser}. As interest in explainable AI has increased, so has interest in evaluating the quality of explanatory rationales. 
However, this is a challenging task because it can be difficult to pin down exactly what constitutes ``good'' rationales for model predictions \citep{jain2019attention,wiegreffe2019attention,serrano2019attention}. 

Two main strategies that have been proposed in recent work are: 1) to view human-generated rationales as a gold standard and evaluate model-generated rationales in comparison to them; and 2) to 
assess the 
``fidelity'' of a rationale to a prediction using automatic metrics.

The human-gold-standard approach views rationales as an additional form of 
label
that can be collected alongside document-level labels. Because 
NLP tasks tend to involve human-generated labels, it makes intuitive sense that human-generated rationales 
might be considered authoritative. 

When human rationales are not available,
evaluations of machine rationales turn to automatic metrics.
These metrics divorce rationale evaluation from an external standard, seeking instead to judge whether rationales are coherent relative to model behavior. 
Popular recent metrics are {\em sufficiency} and {\em comprehensiveness} (i.e., necessity), which assess whether a rationale is sufficient/necessary for a model prediction by comparing the model's behavior on the full input to its behavior on input masked according
to the rationale or its complement.
We use the term {\it fidelity} to refer jointly to sufficiency and comprehensiveness.

\begin{table*}[]
\small
\begin{tabular}{@{}p{.005\textwidth}p{.49\textwidth}p{.08\textwidth}p{.1\textwidth}p{.1\textwidth}p{.1\textwidth}@{}}
\toprule
   & \multicolumn{1}{c}{\textbf{Human rationale}}                                                                                                                                                                            & \textbf{\textbf{Sufficiency}} & \textbf{\textbf{Comprehen-siveness}} & \textbf{\textbf{Failure type}} & \textbf{Dataset} \\ \midrule
1. & No Way Out is the debut studio album by ... Puff Daddy . \highlightrationale{It was released on July 1 , 1997 , by his Bad Boy record label} . ... [SEP] \highlightrationale{1997 was the year No Way Out was released.} & \problematic{0.005}           & 0.224                                & Human                          & \fever            \\
2. & \highlightrationale{== what the f*** is your problem , b**** !!!!!!!!!!!} == why the f*** did you delete the dreamtime festival page , s******                                                                      & 1.0                           & \problematic{0.001}                  & Human                          & \wikiattack        \\ \cmidrule(l){2-6} 
3. & A man \highlightrationale{sits} on a couch beside a colorful cushion with a pencil in his hand. [SEP] The man is \highlightrationale{laying} down on the couch.                                                         & \problematic{0.002}           & 0.999                                & Metric                         & \esnli            \\
4. & :: makes sense . have a good one .                                                                                                                                                                                      & \problematic{0.971}           & \problematic{0.0}                    & Metric                         & \wikiattack        \\ \bottomrule
\end{tabular}
\caption{Example rationales drawn from various datasets. Underlined tokens are rationales provided by humans.
Human annotators can fail to produce faithful rationales (row 1 and 2), and fidelity metrics themselves can be misleading (row 3 and 4).
}
\label{tab:example}

\end{table*}

To the best of our knowledge, no existing work has systematically examined human rationales using 
these automatic metrics. 
However, this is an important step towards evaluating rationales because it helps characterize the disparities between the two types of approach. Are human rationales sufficient to allow models to predict human labels? Are they comprehensive? And what other insights can we gain about human rationales and fidelity metrics by performing this assessment?

In practice, both human rationales and automatic metrics can fail to work as intended (Table \ref{tab:example}).
For instance, human rationales may be insufficient because they fail to include needed information (e.g., the album title in Table \ref{tab:example}.1), or non-comprehensive because they miss redundant-yet-relevant information (e.g., the second personal attack in Table \ref{tab:example}.2).

By contrast, a truly sufficient rationale can be deemed insufficient due to a model not learning expected classification rules (e.g., ``sits'' $\sim$ ``laying'' in Table \ref{tab:example}.3). While this type of failure is inevitable in machine learning, more avoidable are cases where model bias causes rationales to be evaluated incorrectly or inconsistently. 
For instance, in Table \ref{tab:example}.4, the model has learned a heavy bias toward the \nontoxic class (i.e., the model predicts \nontoxic for the empty input), so an empty rationale for a \nontoxic prediction is deemed  perfectly sufficient yet entirely noncomprehensive. 

To investigate the empirical properties of human rationales and automatic metrics,
we analyze the fidelity of human rationales across six datasets.
We show that human rationales do not necessarily have 
high sufficiency or comprehensiveness based on automatic metrics, and 
their fidelity varies greatly from model to model and class to class.
We propose 
extensions to
existing fidelity metrics and develop novel methods to further characterize the quality of human rationales.
First, we note that fidelity is highly model-dependent, and that model behavior can result in misleading fidelity results.
We propose a normalization procedure to allow for fair comparison of these metrics across models, classes, and datasets. We show that this normalization helps contextualize fidelity results by accounting for baseline model behavior.

Second, we evaluate model accuracy on full vs. rationale-only data, linking typical output-sufficiency to performance outcomes (i.e., {\em accuracy-sufficiency}).
We examine the effect of allowing models to adapt to 
rationale-only data
during training, drawing a distinction between a rationale's ``incidental'' fidelity and its ``potential'' fidelity to a model. 
We analyze the effect of these two interventions and discuss their implications for evaluation of (and learning from) human rationales. 

Finally, we introduce the idea of ``fidelity curves'', which examine how sufficiency and comprehensiveness degrade as tokens are randomly occluded from a rationale. We discuss how the shapes of these curves can be used to infer fine-grained attributes about rationales, such as the extent to which they contain redundant or highly interdependent tokens. 
We find that rationales in our datasets vary greatly in their level of irrelevancy, redundancy, and mutual dependence. We find that our three 
classification tasks exhibit 
less dependence and more redundancy in their rationales
than our three \qastyle 
tasks.

Evaluating rationales is a 
significant
challenge. We argue that in order to be confident in either 
human rationales or automatic fidelity metrics, we have to understand how these two evaluation approaches interact with one another, and what caveats they can reveal about each other. 
Our analyses lead to the following actionable implications:

\begin{itemize}[itemsep=0pt,leftmargin=*,topsep=0pt]
  \item Fidelity metrics are highly model-dependent and should be normalized to assist interpretation.
  \item Models trained on rationale-only data can provide accuracy-based metrics to complement the ``incidental'' metrics. 
  \item 
  ``Fidelity curves'' provide a novel way to infer fine-grained qualities about rationales, such as irrelevance and redundancy. 
\end{itemize}

\section{Datasets}
\label{sec:datasets}

\begin{table*}[]
\centering
\small
\begin{tabular}{llllllll}
\hline
\multirow{2}{*}{Dataset} & \multirow{2}{*}{\begin{tabular}[c]{@{}l@{}}Text\\ length\end{tabular}} & \multirow{2}{*}{\begin{tabular}[c]{@{}l@{}}Task type\end{tabular}} & \multicolumn{5}{c}{Rationale}                                                                                                                         \\ \cline{4-8} 
                         &                                                                        &                                                                      & Length & Ratio  & \begin{tabular}[c]{@{}l@{}}Comprehensive\end{tabular} & Granularity & \begin{tabular}[c]{@{}l@{}}Class\\ asymmetry\end{tabular} \\ \hline

 WikiAttack &         51.8 &       classification &               6.5 &    19.1\% &     \cmark &       Token &       \cmark \\
        SST &         19.3 &       classification &               6.5 &    34.6\% &     \cmark &       Token &       \xmark \\
      Movie &        774.3 &       classification &              82.4 &    11.3\% &     \xmark &       Token &       \xmark \\
    MultiRC &        321.7 &        \qastyle &              69.8 &    22.9\% &     \cmark &    Sentence &       \xmark \\
      FEVER &        320.7 &        \qastyle &              53.6 &    24.0\% &     \xmark &    Sentence &       \xmark \\
     E-SNLI &         21.3 &        \qastyle &               5.0 &    25.2\% &     \cmark &       Token &       \cmark \\
\hline
\end{tabular}
\caption{Basic statistics. Dataset rationales exhibit a range of average rationale-to-text ratios, expected comprehensivenesses, granularities, and class asymmetries.}
\label{tab:statistics}
\end{table*}

The goal of this paper is to 
evaluate and characterize human rationales.  
We analyze six datasets, four drawn from the ERASER collection \citep{deyoung2019eraser}, and two from other sources. They consist of three single-text classification tasks and three \qastyle tasks where it is important to understand the relations between texts.

For each dataset, the human rationales have a qualitative \textit{expected comprehensiveness} based on whether, by construction or design, they are intended to contain all pertinent information for their respective prediction task. Four of our six datasets are expected to have comprehensive rationales.

\begin{itemize}[itemsep=0pt,leftmargin=*,topsep=-2pt]
\item \textbf{\wikiattack} \citep{carton_extractive_2018}. A classification dataset of 115,859 Wikipedia revision comments labeled for presence of personal attacks by \citet{wulczyn_ex_2017} and augmented with 1,049 human rationales by \citet{carton_extractive_2018}.
The rationales in this dataset are expected to be comprehensive, as labelers were asked to identify all personal attacks in each text. 

\item \textbf{Stanford Sentiment Treebank (SST)} \citep{socher2013recursive}.
A classification dataset of 9,620 movie review snippets annotated for positive/negative sentiment at every syntactic tree node. We flatten these into rationales using a heuristic algorithm (see the appendix). The rationales are expected to be comprehensive, as they contain all high-sentiment phrases.

\item \textbf{\movies} \cite{zaidan2008modeling}. A classification dataset of 2,000 movie reviews labeled with rationales. The rationales 
are not necessarily comprehensive, as annotators were not instructed  to identify all evidence.

\item \textbf{MultiRC} \citep{khashabi2018looking} A reading comprehension dataset of 32,091 document-question-answer triplets that are true or false.
Rationales are expected to be comprehensive as they each consist of 2-4 sentences from a document that are required to answer the given question.

\item \textbf{FEVER} \cite{thorne2018fever} A fact verification dataset of 76,051 snippets of Wikipedia articles paired with 
claims that they support or refute. Rationales consist of a single contiguous sub-snippet (and the claim itself), and are not expected to be comprehensive as they may not cover all pertinent information.

\item \textbf{\esnli} \cite{camburu2018snli} A textual entailment dataset of 568,939 short snippets and claims for which each snippet either refutes, supports, or is neutral toward. Explanations for this dataset are expected to be comprehensive as the texts are short and labelers were instructed to identify all relevant tokens.
\end{itemize}

\begin{figure}[t]
  \centering
  \includegraphics[width=0.7\linewidth]{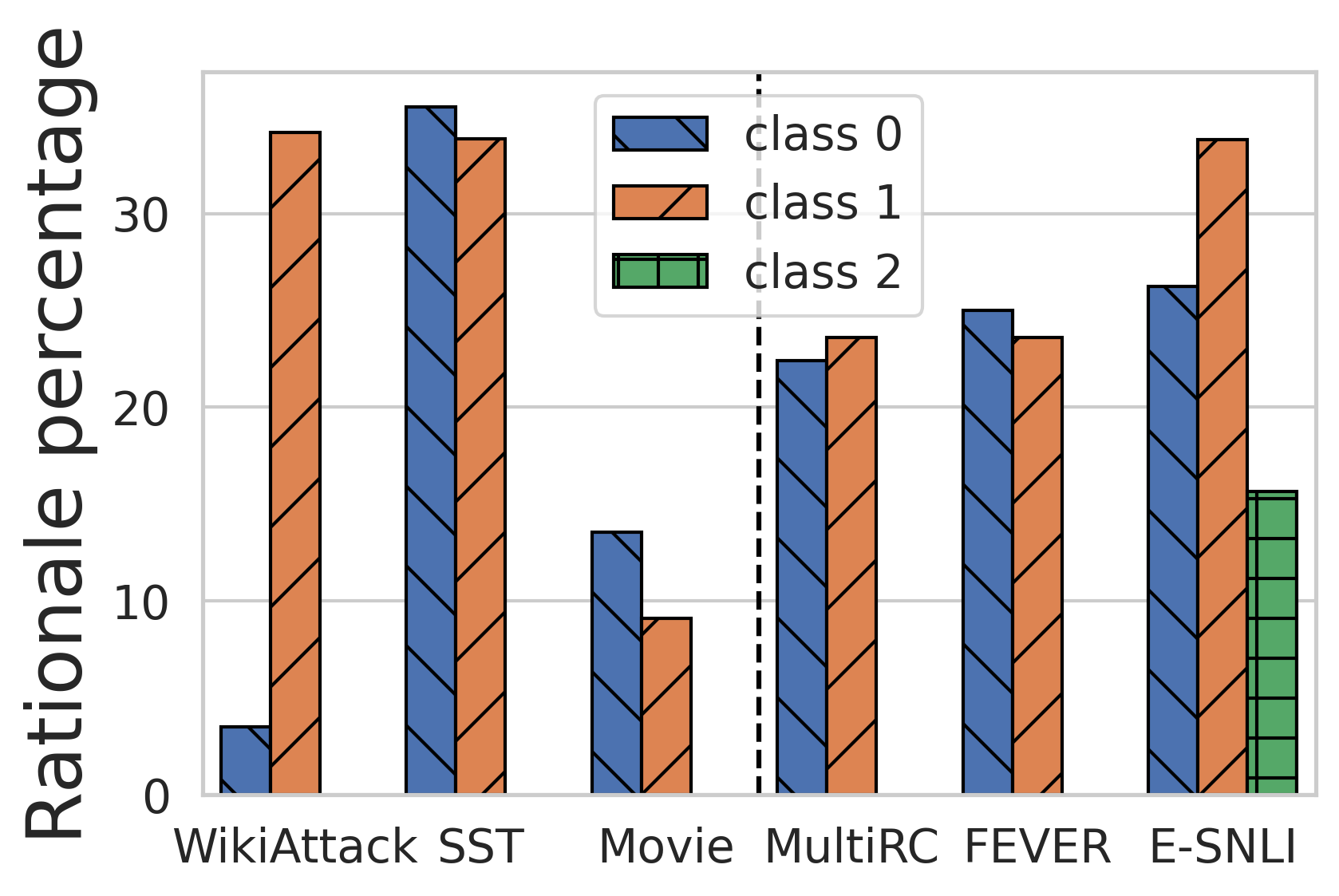}
  \small
  \begin{tabular}{lp{0.7\linewidth}}
  \toprule
   \wikiattack   & 0: \nontoxic, 1: \toxic \\
   \sst       & 0: negative, 1: positive\\
   \movie       & 0: negative, 1: positive\\
   \multirc       & 0: false, 1: true\\
   \fever       & 0: refutes, 1: supports\\
   \esnli & 0: contradiction, 1: entailment, 2: neutral\\
  \bottomrule
  \end{tabular}
  \caption{Percentage of rationales by class. 
  Significant variations exist 
  in \wikiattack and \esnli.
  }
  \label{fig:rationale_per}
\end{figure}

\noindent\tableref{tab:statistics} shows the basic statistics of each dataset.
Significant variation exists between datasets in rationale length and rationale percentage.
For example, rationales only cover 11.3\% of the words in \movie, consistent with our expectation of non-comprehensiveness. We also report rationale granularity, whether annotations were provided at the token or sentence level, and class asymmetry, whether rationale lengths vary significantly between classes. For the purpose of this analysis, tokenization is provided by the individual dataset sources, so we simply split texts by whitespace.

\figref{fig:rationale_per} shows class asymmetry in rationale percentages.
For \wikiattack, labelers were asked to highlight personal attacks, and thus evidence for the \nontoxic class comes in the form of no highlighted tokens. This results in a situation where 
rationales for \nontoxic examples constitute less than 5\% on average, while they constitute 35\% of \toxic examples. 
Significant variation between classes also exists in \esnli: entailment 
contains close to 40\% of tokens as rationales, but neutral merely consists of 16\% --- another case of evidence through absence (negative evidence).

\section{Evaluating Human Rationales}
\label{sec:evaluation}

Popular automatic metrics for evaluating machine-generated rationales are {\em sufficiency} and {\em comprehensiveness}, articulated by \citet{yu_rethinking_2019} and employed in the ERASER benchmark \citep{deyoung2019eraser}.
{\em Sufficiency} measures how well rationales can provide the same prediction as using full information,
while {\em comprehensiveness} measures how well rationales include all relevant information.

It remains an open question whether human-generated rationales have good sufficiency and comprehensiveness.
We find that this is in fact not necessarily the case.
This result reveals a contradiction in the evaluation of machine-generated rationales:
human-generated rationales are used as a gold standard, but being similar to human-generated rationales may not lead to high sufficiency and comprehensiveness.
Another important observation from our experiments is that there exists significant variation between datasets and classes within the same dataset.

\begin{figure*}[t]
\centering
\begin{subfigure}[t]{0.30\textwidth}
  \centering
    \includegraphics[width=\linewidth]{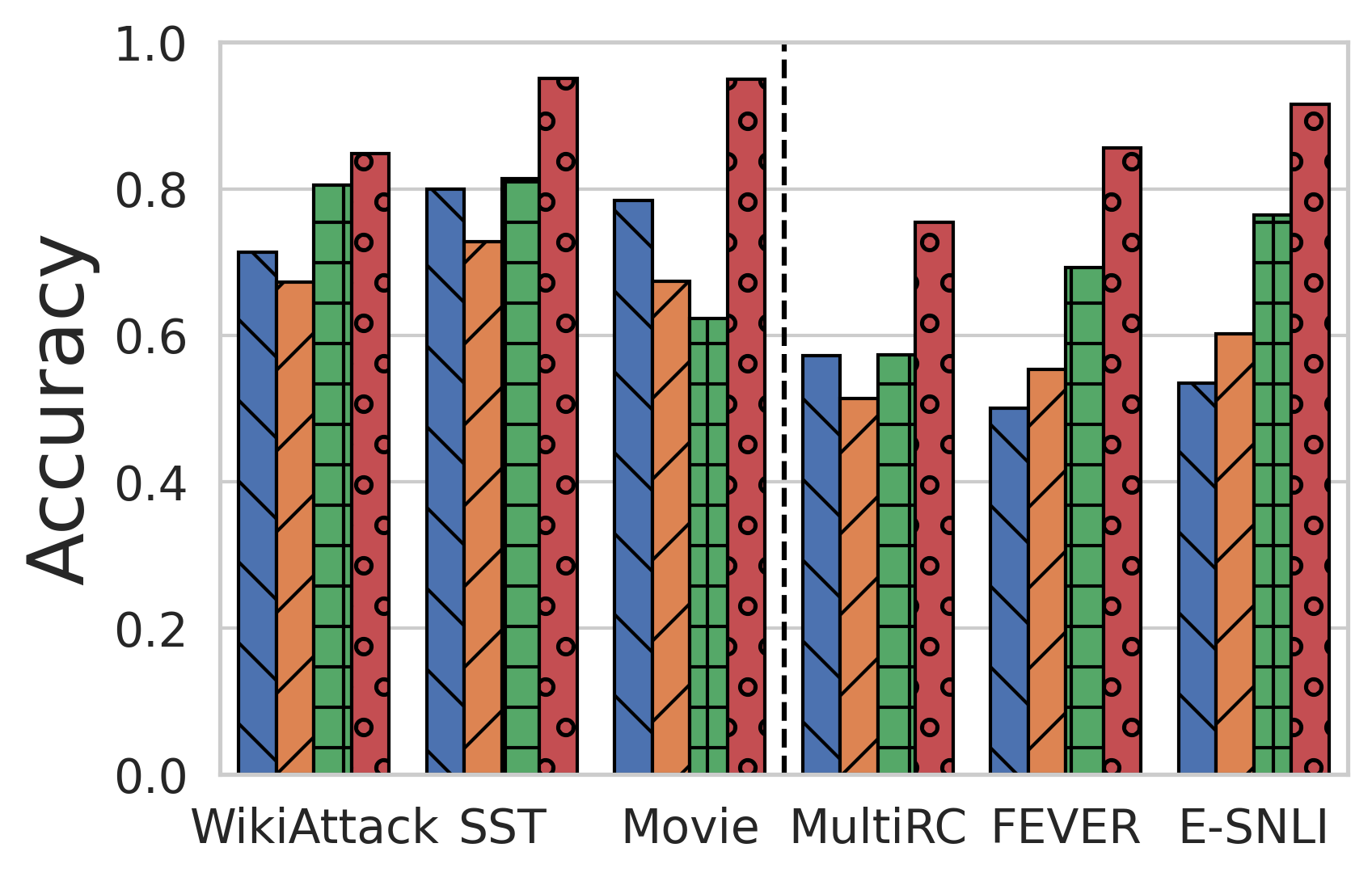}
  \caption{Accuracy}
  \label{fig:overall_accuracy}
\end{subfigure}
\begin{subfigure}[t]{0.30\textwidth}
  \centering
      \includegraphics[width=\linewidth]{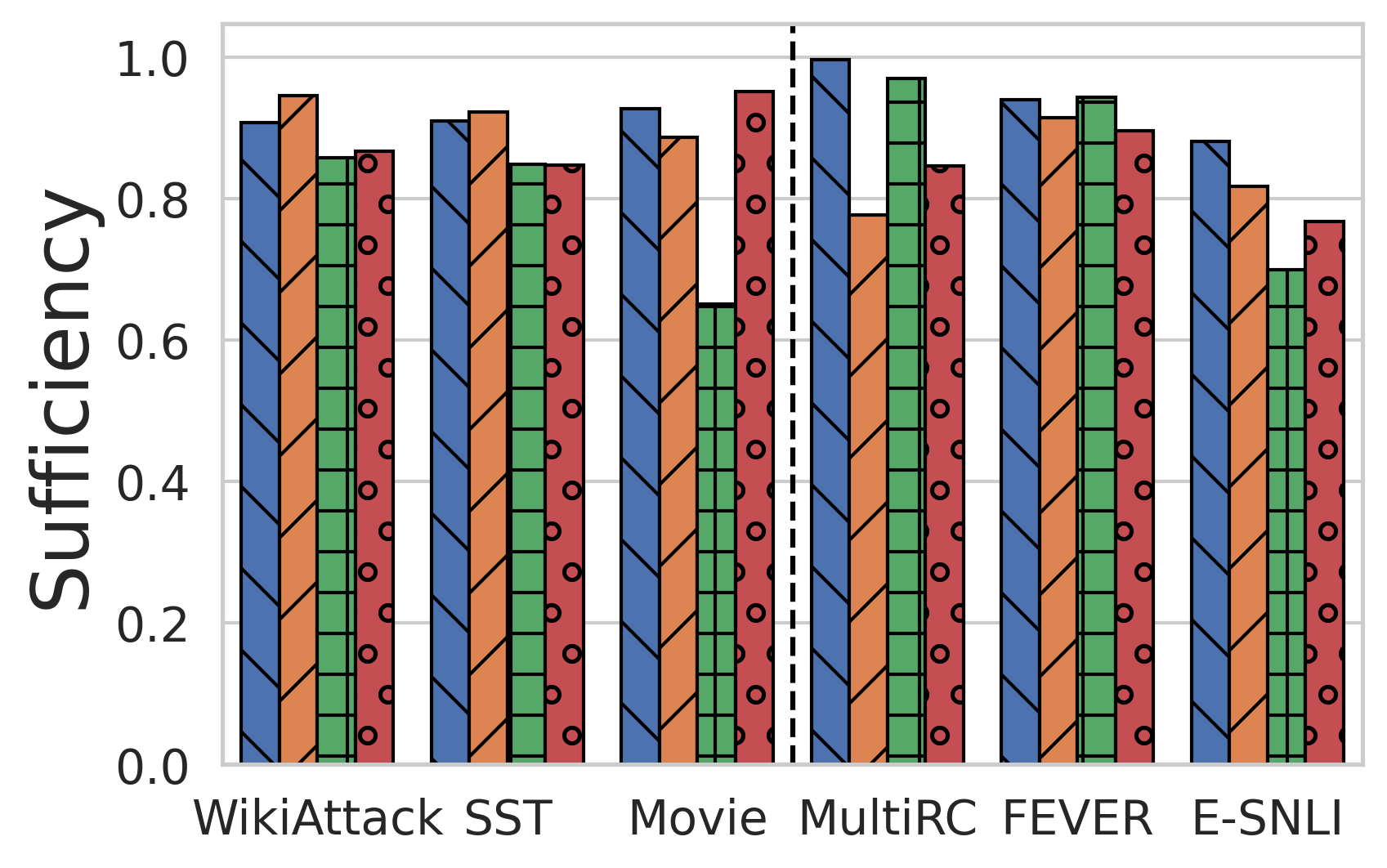}

  \caption{Sufficiency}
  \label{fig:overall_sufficiency}
\end{subfigure}
\begin{subfigure}[t]{0.30\textwidth}
  \centering
    \includegraphics[width=\linewidth]{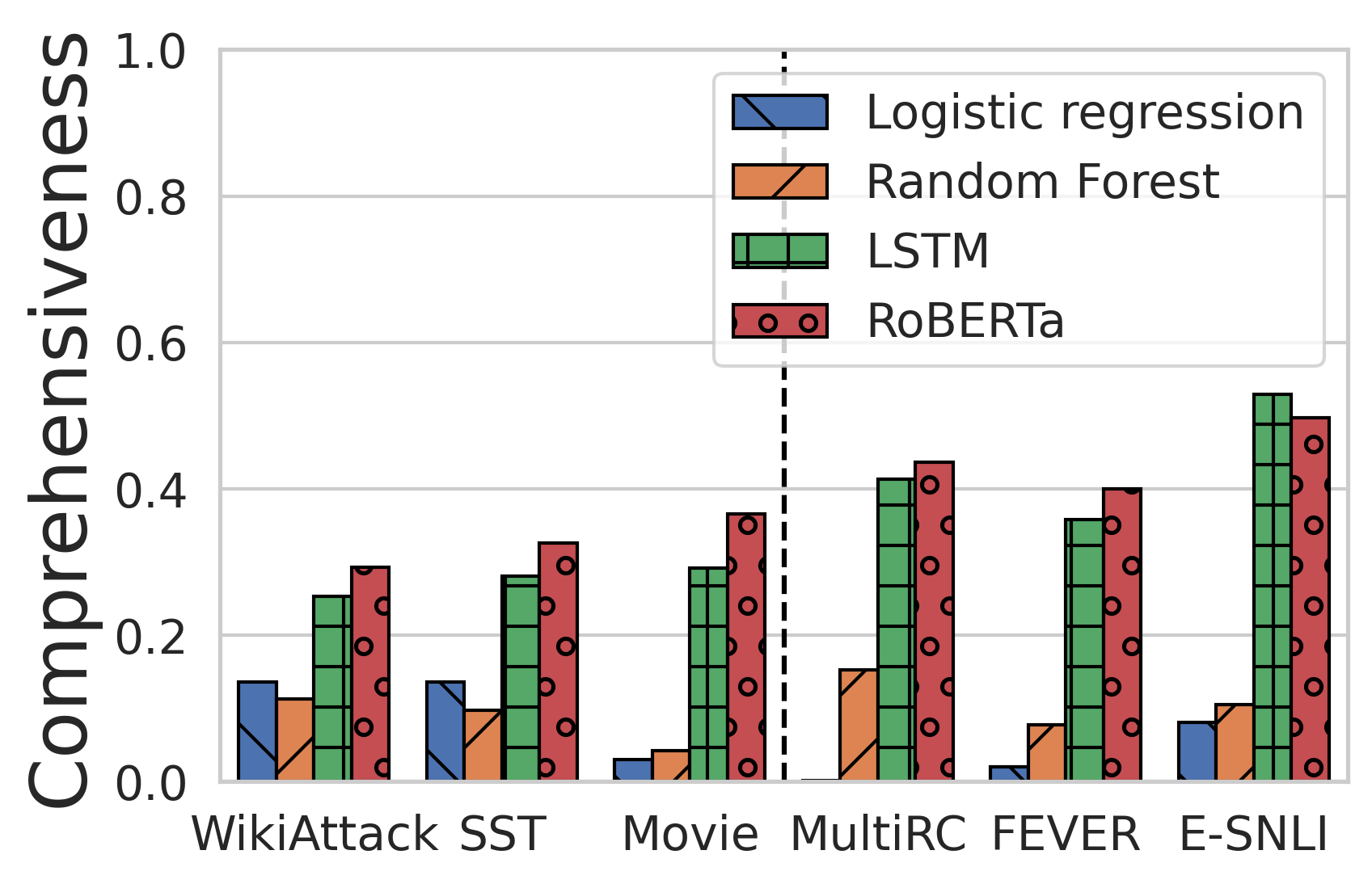}

  \caption{Comprehensiveness}
  \label{fig:overall_comprehensiveness}
\end{subfigure}
\caption{Accuracy, sufficiency, and comprehensiveness of human rationales with different models.
While \roberta performs significantly better in all datasets in accuracy, it is rarely the best in sufficiency.
In comparison, human rationales tend to have abysmal comprehensiveness with classic models.
}
\label{fig:overall}
\end{figure*}

\begin{figure*}[t]
  \centering
    \begin{subfigure}[t]{0.30\textwidth}
        \includegraphics[width=\linewidth]{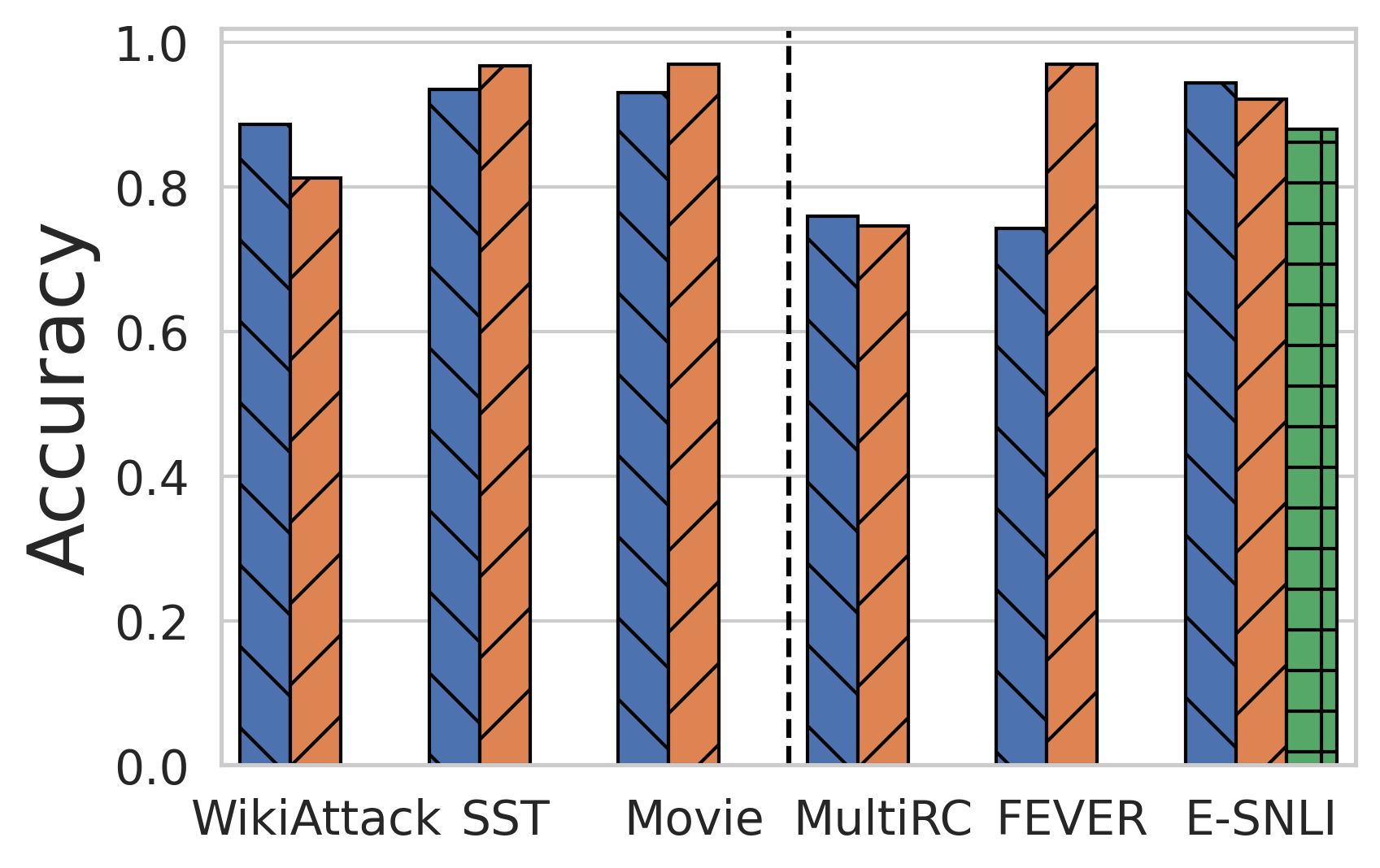}  
        \caption{Accuracy by class}
        \label{fig:accuracy_by_class}
    \end{subfigure}
    \begin{subfigure}[t]{0.30\textwidth}
        \includegraphics[width=\linewidth]{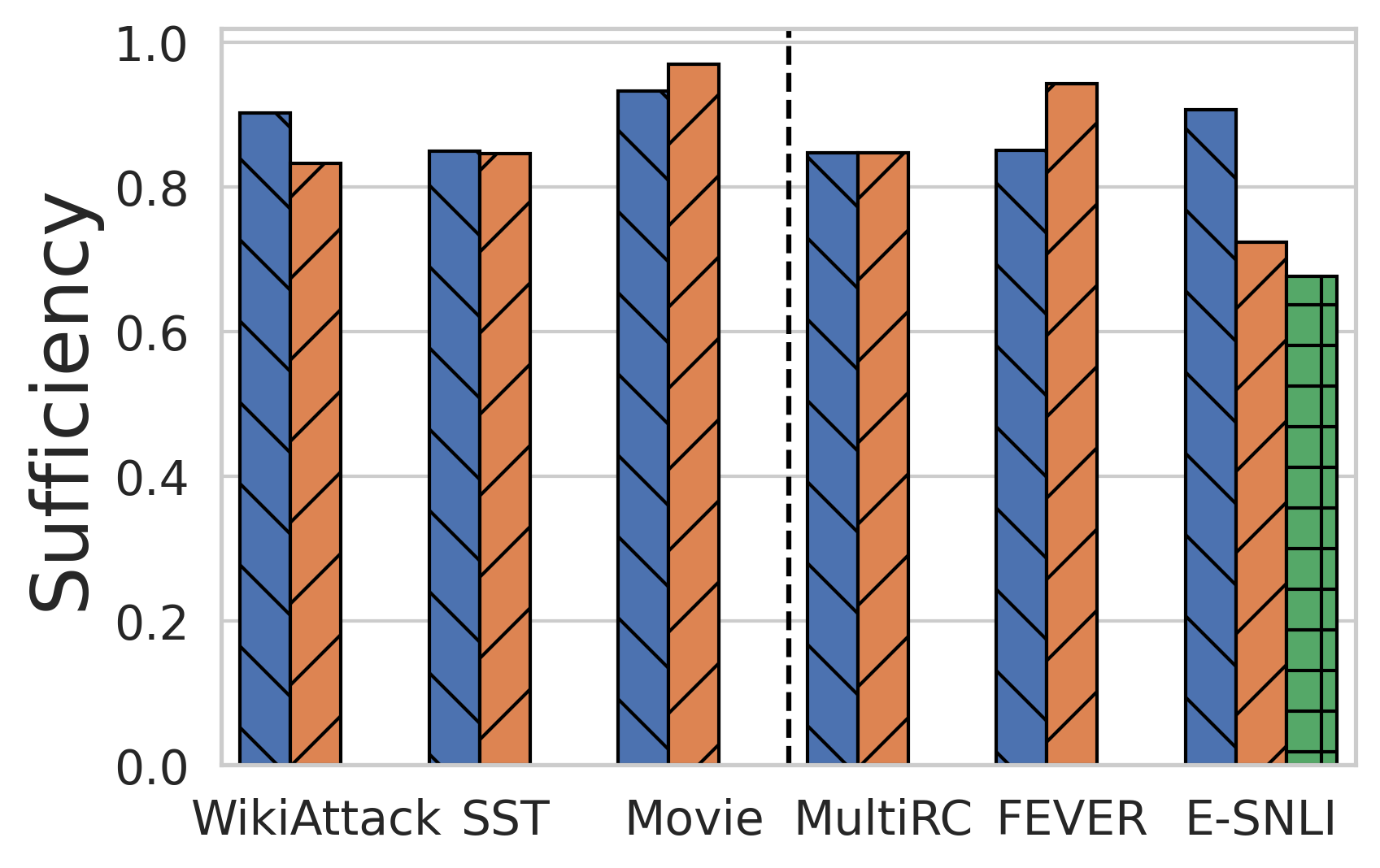}

        \caption{Sufficiency by class}
        \label{fig:sufficiency_by_class}
    \end{subfigure}
    \begin{subfigure}[t]{0.30\textwidth}
        \includegraphics[width=\linewidth]{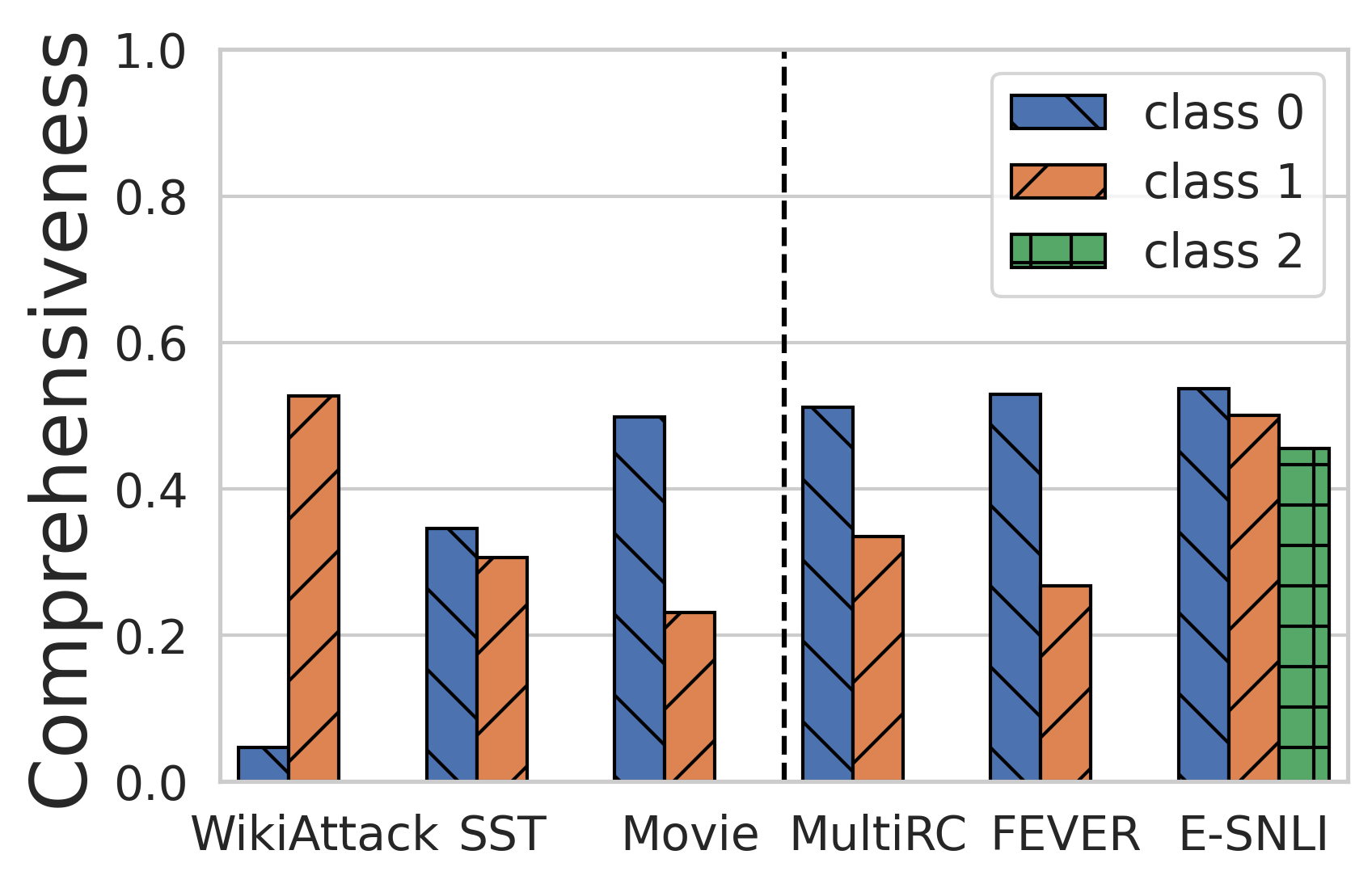}

        \caption{Comprehensiveness by class}
        \label{fig:comprehensiveness_by_class}
    \end{subfigure}
    \caption{Accuracy, sufficiency, and comprehensiveness of human rationales grouped by class for \roberta.
    While sufficiency is relatively stable across classes, we observe dramatic differences between classes in comprehensiveness (e.g., \wikiattack and \movie).
    }
    \label{fig:all_by_class}
\end{figure*}

\subsection{Formal Definitions \& Experiment Setup}

A rationale is \textit{sufficient} if it contains enough information to allow the model to make a prediction close to what it would make with full information. 
Formally, we represent rationales as a binary mask $\rationale$ over the input $\data$ that indicates whether each token belongs to the rationale or not (1 to include, 0 to exclude).
The sufficiency of rationales for a given prediction $\hat{y}$ is 
based on the difference in class probability 
between
using full information and
using only the rationale: 
\begin{equation}
    \small
    \mathrm{Suff}(\data,\hat{y}, \rationale) = 1-\max(0, p(\hat{y}|\data) - p(\hat{y}|\data, \rationale)),
\end{equation}
where $\hat{y} = \argmax_y p(y|\data)$.
Note that we use the reverse of the difference so that higher sufficiency indicates faithful rationales.
We also 
enforce the difference in class probability to be above 0, which differs from \citet{deyoung2019eraser}.\footnote{Arguably, the sufficiency metric should not go above 1 no matter how good the rationales are. 
That said, our results demonstrate similar qualitative trends from the definitions without the max operation. See the appendix.}
This operation bounds sufficiency to between 0 and 1.

Comprehensiveness (i.e., necessity) captures the extent to which a rationale is needed for a prediction, by assessing the model's prediction on the complement of the rationale (${\bm 1}-\rationale$). For a highly comprehensive explanation, the model's prediction on its complement should differ greatly from its prediction on the full information. 
As above, we enforce this value to be bounded between 0 and 1:
\begin{equation}
\small
\mathrm{Comp}(\data,\hat{y}, \rationale) = \max(0, p(\hat{y}|\data) - p(\hat{y}|\data, {\bm 1}-\rationale)).    
\end{equation}
Our definitions entail that a faithful rationale should have both 
high sufficiency and comprehensiveness.

Implicit in the definition of sufficiency and comprehensiveness is a dependence on the properties of the underlying model.
To study the relationship between model property and human rationale fidelity, we experiment with a range of models: logistic regression, random forests, LSTM \citep{hochreiter1997long} and \roberta \citep{liu_roberta:_2019}.
We use the same train/dev/test splits as in the original datasets.
We report the resulting model with the best validation accuracy in the main paper.
To apply rationale masking, 
we simply remove the tokens which correspond with 0s in the rationale mask.
See the supplementary material for implementation details.
Our code is available at \url{https://github.com/BoulderDS/evaluating-human-rationales}.

\subsection{Overall Results}

\begin{figure*}[t]
\centering
  \begin{subfigure}[t]{0.3\textwidth}
    \includegraphics[width=\linewidth]{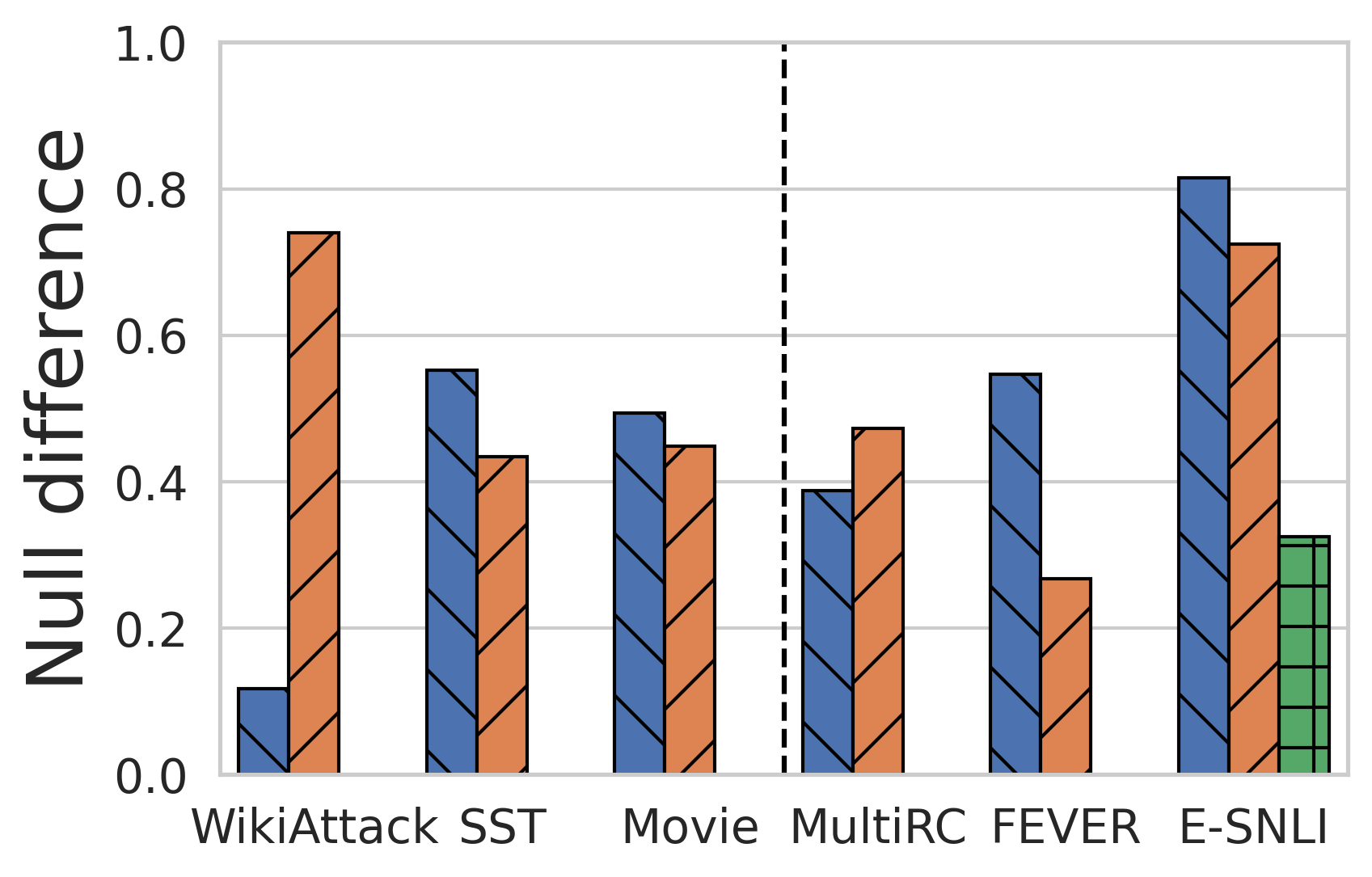}

    \caption{Null difference. %
    }
    \label{fig:null_bert_fidelity}
   \end{subfigure}
  \centering
  \begin{subfigure}[t]{0.3\textwidth}
    \includegraphics[width=\linewidth]{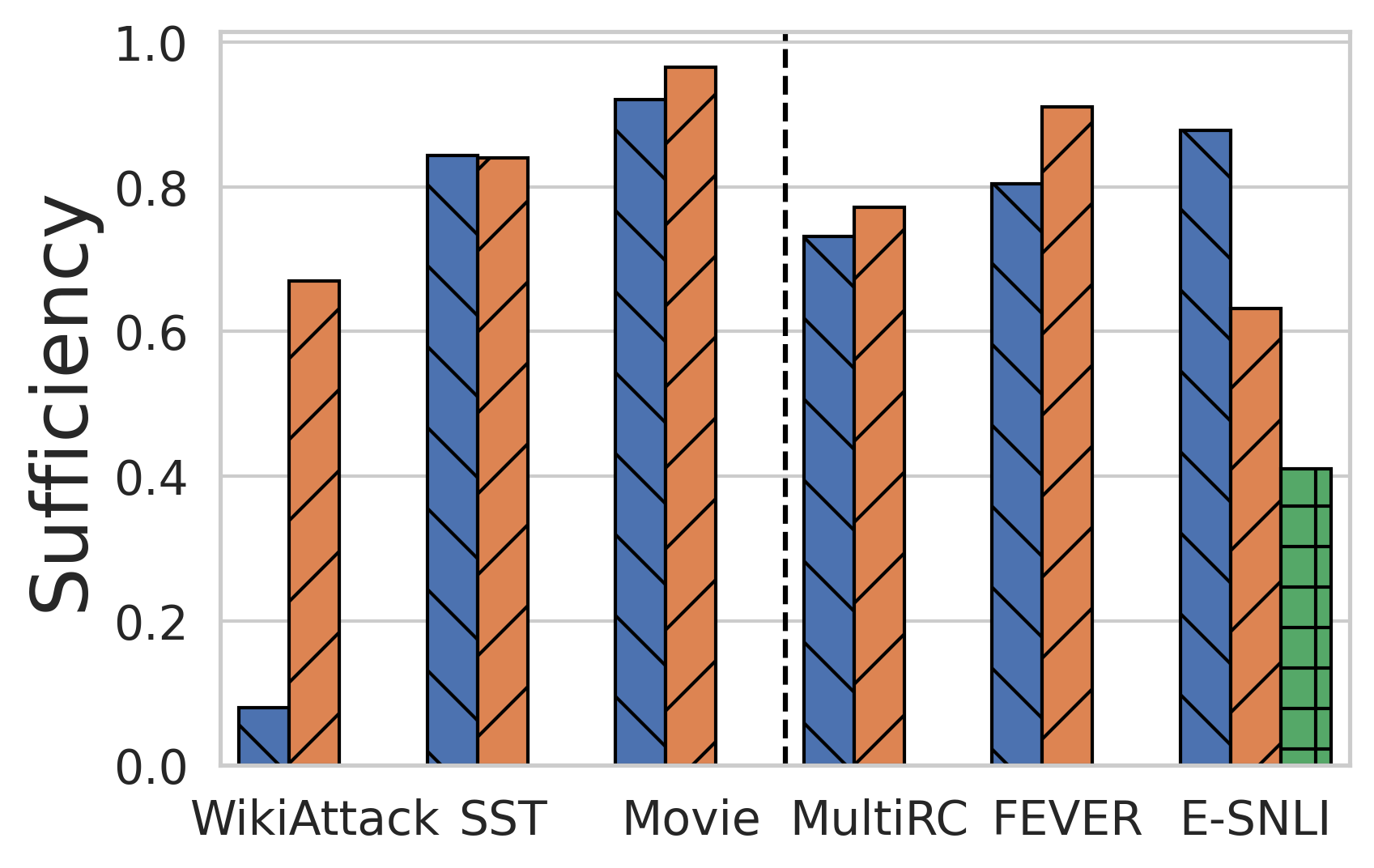}

    \caption{Normalized sufficiency.
    }
    \label{fig:normalized_bert_sufficiency_by_class}
   \end{subfigure}
    \begin{subfigure}[t]{0.3\textwidth}
    \centering

    \includegraphics[width=\linewidth]{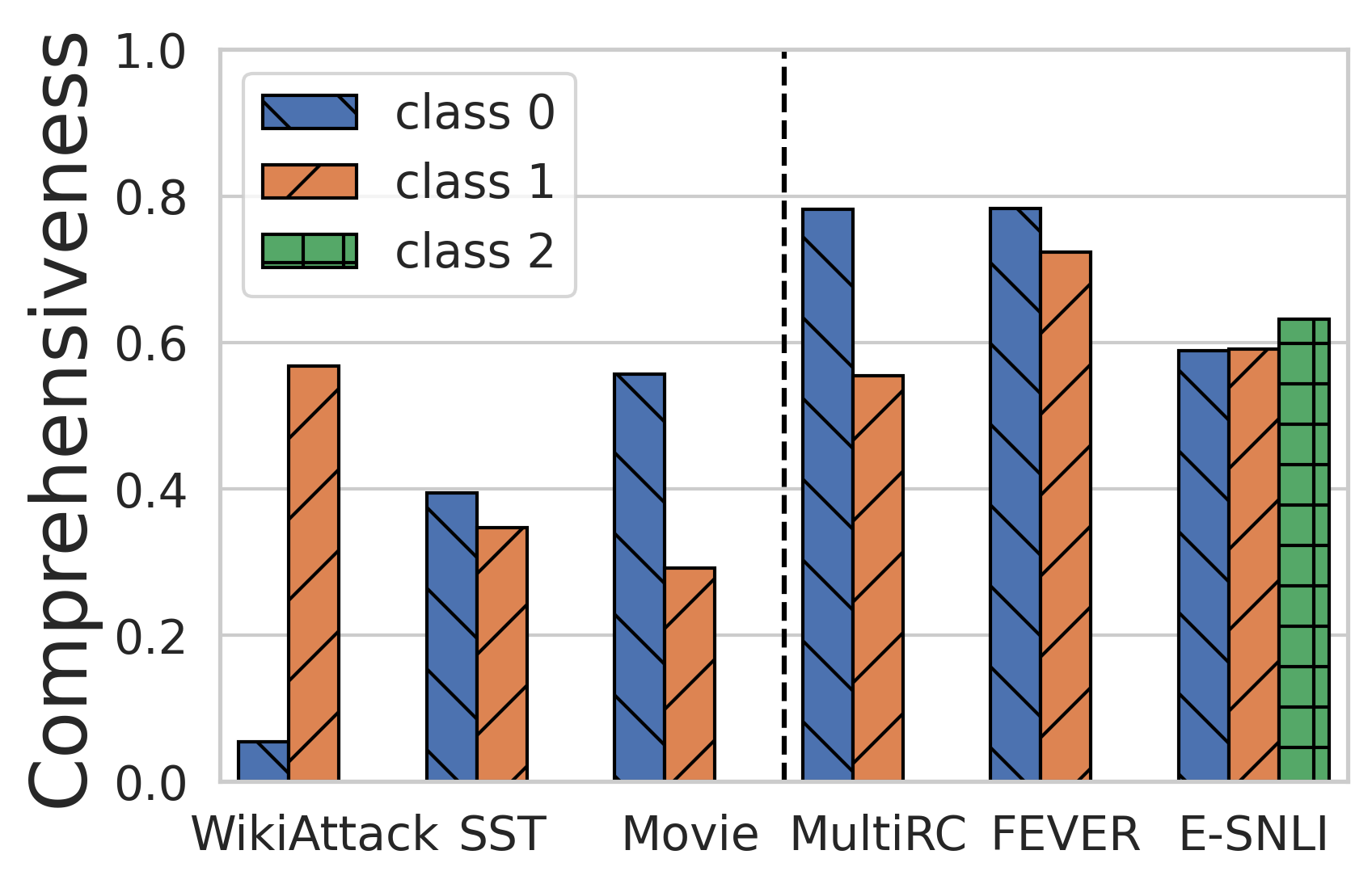}

  \caption{Normalized comprehensiveness.}
  \label{fig:normalized_bert_comprehensiveness_by_class}
   \end{subfigure}
   \caption{Normalization is critical for interpreting sufficiency and comprehensiveness. Here we show evaluations of human-generated rationales based on \roberta.}
   \label{fig:normalized_bert}
\end{figure*}

\figref{fig:overall_accuracy} shows the accuracy of our models on each dataset. As expected, \roberta shows the best performance followed generally by LSTM, then random forest and logistic regression. 
The only exception is \movie, where LSTM models struggle with the long texts (774 tokens on average) due to 
the limited dataset size and vanishing gradients.

We find that {\bf human rationales do not necessarily have high sufficiency and comprehensiveness}.
Moreover, human-generated rationales obtain weaker sufficiency in highly accurate models (\figref{fig:overall_sufficiency}).
In fact, human rationales have lower sufficiency in \roberta than logistic regression or random forest in five of six datasets.
This finding demonstrates that the sufficiency of an explanation can be inversely correlated with model performance, which is a problem for comparing explanation methods across different models.
By contrast, strong models show better comprehensiveness scores for human rationales (\figref{fig:overall_comprehensiveness}), with values ranging from 0.3 to 0.5 for \roberta.
\esnli demonstrates the highest comprehensiveness in this model while \movie and \multirc, both expected to be non-comprehensive, respectively achieve the 2nd and 4th highest comprehensiveness, in defiance of our expectations.

Moving forward, we focus on \roberta 
as it is the most accurate and represents the industry standard for general NLP. 

\para{Classes matter.}
Breaking down fidelity by class reveals further nuances.
\figref{fig:sufficiency_by_class} shows that sufficiency is mostly even between classes, though significant differences exist for \esnli. Surprisingly, in \wikiattack, sufficiency is higher in the \nontoxic class where there are a small number of tokens in human rationales.

The evenness in sufficiency
is not mirrored in
comprehensiveness (Fig. \ref{fig:comprehensiveness_by_class}), 
which differs wildly from class to class for different datasets. The most extreme case is \wikiattack, where by design the ``rationale'' for a \nontoxic comment is for nothing to be highlighted. The comprehensiveness of these empty rationales is 
correspondingly
 low. Interestingly, \esnli demonstrates a relatively even spread of comprehensiveness across classes despite its class-asymmetric rationale lengths.

\movies, \multirc, and \fever all show large class discrepancies in comprehensiveness despite having 
similar-length rationales across classes. In \fever, for example, this means that removing the identified evidence for a ``refutes'' outcome tends to have a higher impact on the model prediction than for ``support'' outcomes. This could be due to task semantics (e.g., that refuting evidence is generally more unique than supporting evidence), or model bias (e.g., that the model tends to predict ``supports'' by default and therefore is less affected by removing the rationales for this outcome).

\section{Normalizing Sufficiency and Comprehensiveness}
\label{sec:metrics}

Human rationales do not necessarily have high 
fidelity, suggesting that either human rationales or evaluation metrics may be problematic.
We start by rethinking the 
fidelity metrics in this section and will propose novel methods to characterize human rationales in \secref{sec:characterizing_beyond}.

A salient observation in \figref{fig:overall} is that sufficiency and comprehensiveness are in completely separate value ranges, although they are both theoretically bounded between 0 and 1.
To properly interpret these numbers, we need to establish a baseline for them.
We do so by 
defining a 
``{\em null difference}'', the difference in output between the model operating on full information vs. no information (i.e., the empty input). 
This value is equivalent to (the complement of) the sufficiency of an all-zero (empty) rationale mask, or the comprehensiveness of an all-one mask. 

Null difference is an intrinsic value for a given model and dataset, and depends on the class balance of the dataset, the bias term(s) learned by the model, and the calibration of output probability. It serves as a baseline value
in the sense that no rationale should be much less sufficient than an all-zero rationale
or much more comprehensive than an all-one rationales.
By normalizing sufficiency and comprehensiveness scores against this value, we can estimate
how faithful rationales are relative to the baseline fidelity of the model.

We use min-max normalization to normalize sufficiency and comprehensiveness with this null difference.
Formally, we define the metrics as follows:
\begin{eqnarray}
& \mbox{ \small $\mathrm{NullDiff}(\data, \hat{y}) 
 =  \max(0, p(\hat{y}|\data) - p(\hat{y}|\data, {\bm 0}))$} \label{eq:null}\\
& \mbox{\small $\mathrm{NormSuff}(\data, \hat{y}, \rationale) = \frac{\operatorname{Suff}(\data, \hat{y}, \rationale) - \operatorname{Suff}(\data, \hat{y}, {\bm 0})}{1 - \operatorname{Suff}(\data, \hat{y}, {\bm 0})}$} \label{eq:normsuff} \\
\small
& \mbox{\small $\mathrm{NormComp}(\data, \hat{y}, \rationale) = \frac{\operatorname{Comp}(\data, \hat{y}, \rationale)}{\operatorname{Comp}(\data, \hat{y}, {\bm 1})}$} \label{eq:normsuff}
\end{eqnarray}
where \mbox{\small $\hat{y}=\argmax_y p(y|\data)$}.
Note that \mbox{\small $\mathrm{NullDiff}(\data, \hat{y}) = 1 - \operatorname{Suff}(\data, \hat{y}, {\bm 0}) = \operatorname{Comp}(\data, \hat{y}, {\bm 1})$}.
We clip \mbox{\small $\mathrm{NormSuff}$} and \mbox{\small $\mathrm{NormComp}$} between 0 and 1.

\begin{figure*}[t]
\centering
\begin{subfigure}[t]{0.29\textwidth}
  \centering
    \includegraphics[width=0.8\textwidth]{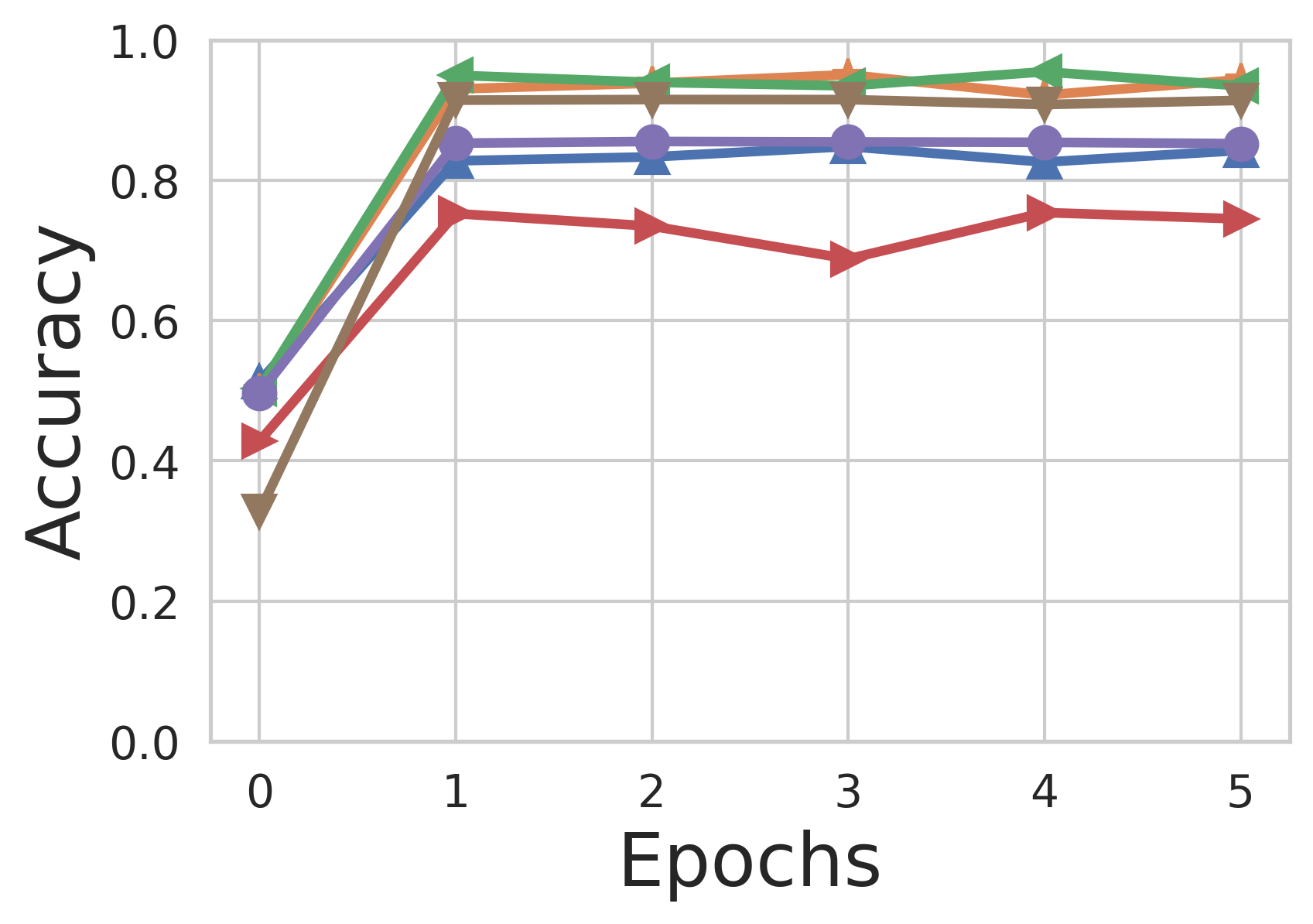}

  \caption{Accuracy}
  \label{fig:Accuracy}
\end{subfigure}
\begin{subfigure}[t]{.29\textwidth}
  \centering
  \includegraphics[width=0.8\textwidth]{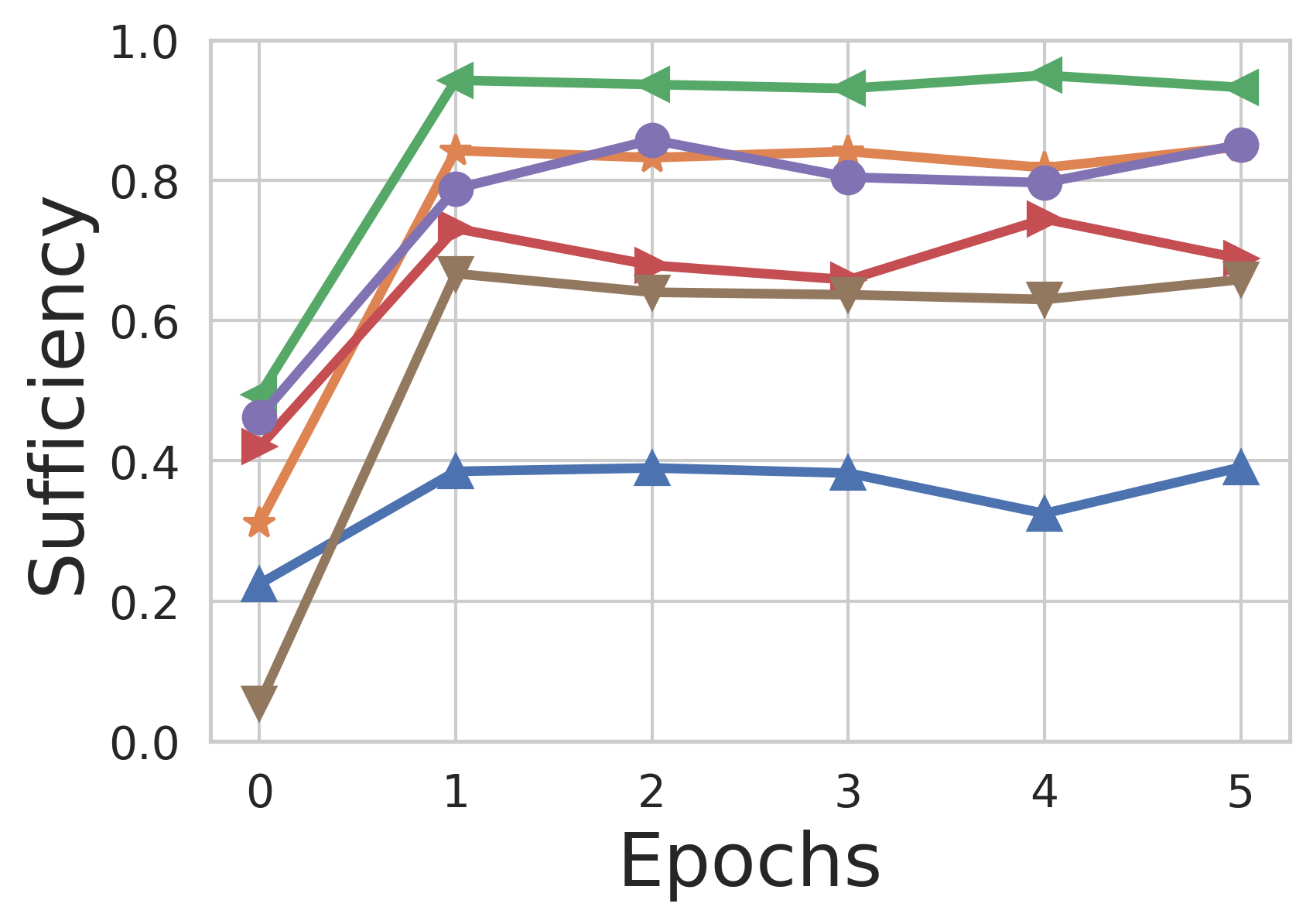}

  \caption{Normalized sufficiency}
  \label{fig:sufficiency}
  \vspace{0.7em}
\end{subfigure}
\begin{subfigure}[t]{0.4\textwidth}
  \centering
    \includegraphics[width=0.8\textwidth]{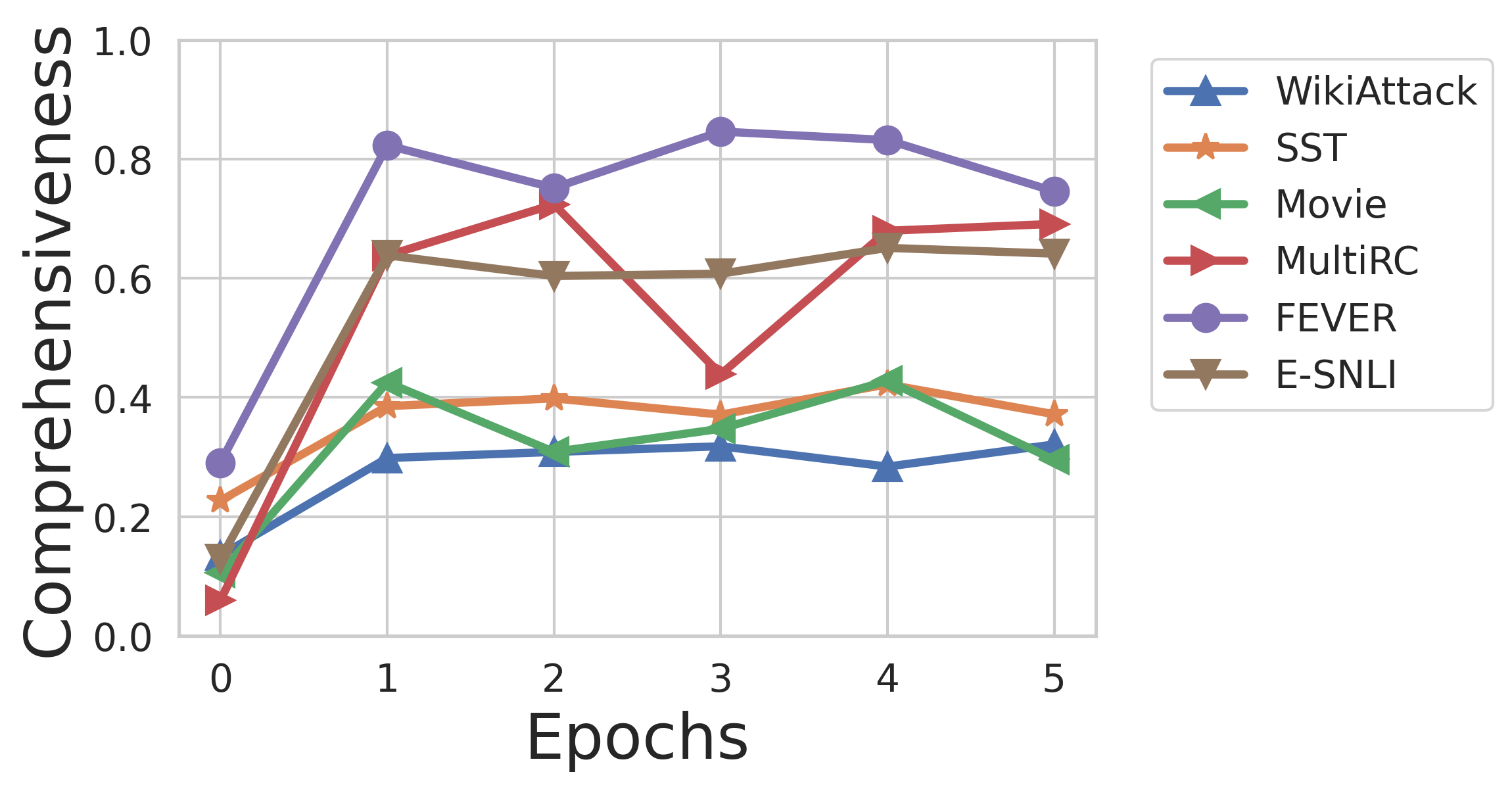}
  \caption{Normalized comprehensiveness}
  \label{fig:comprehensiveness}
\end{subfigure}
\caption{Accuracy, normalized sufficiency, and normalized comprehensiveness vs. \#epochs in \roberta.
While accuracy stabilizes after 1 epoch, sufficiency and comprehensiveness demonstrate significant fluctuation. 
}

\label{fig:bert_epochwise_fidelity}
\end{figure*}

\figref{fig:null_bert_fidelity} shows the null difference for \roberta across all datasets by class. 
Significant variation exists between classes, especially for \wikiattack, \fever, and \esnli, an observation that helps contextualize some of the results in \figref{fig:all_by_class}, as reflected by the normalized fidelity metrics.

\figref{fig:normalized_bert_sufficiency_by_class} shows that normalized sufficiency is much lower in the \nontoxic class in \wikiattack, meaning that \nontoxic rationales 
are barely more informative than an empty rationale. 
This resolves the puzzle that the short/empty rationales in the \nontoxic class have high sufficiency in \figref{fig:sufficiency_by_class}.
It is also more consistent with the low comprehensiveness measured for these rationales.

\figref{fig:normalized_bert_comprehensiveness_by_class} shows us that 
the comprehensiveness scores even out for \fever under this normalization, suggesting that the previous result was simply a product of model bias. 
By contrast, the asymmetric scores for \movie and \multirc shown in \figref{fig:comprehensiveness_by_class} cannot be explained by model bias, indicating that 
the interaction between task semantics and model learning 
may cause rationales to be more comprehensive in the negative class than in the positive class for these datasets. 
Another outcome of normalization is to map sufficiency and comprehensiveness to the same scale.
Comprehensiveness in single-text classification tasks are generally lower than that in \qastyle tasks.

These results suggest that sufficiency and comprehensiveness metrics are highly model-dependent and should not be compared across models without care.

\para{Fidelity and model training.}
Examining how human rationale fidelity changes from epoch to epoch as models train (\figref{fig:bert_epochwise_fidelity}) further demonstrates the model-dependence of these measures. 
Random noise causes the models to have nonzero (but low) fidelity scores at epoch 0. However, we observe that even after accuracy stabilizes, sufficiency and comprehensiveness may continue to fluctuate significantly, e.g., \fever sufficiency.\footnote{We observe similar issues with logistic regression, random forest, and LSTM. See the appendix.} Further, the maximum fidelity may not co-occur with the maximum accuracy (e.g., \multirc comprehensiveness). While most of the fluctuation 
isn't drastic,
these differences could prove decisive in a head-to-head comparison of fidelity scores across different models or rationalization techniques. 
These observations suggest that we need to be cautious before 
claiming definitive fidelity for a given model using these automatic metrics. 

\section{Characterizing Human Rationales beyond Sufficiency/Comprehensiveness}
\label{sec:characterizing_beyond}

Sufficiency and comprehensiveness offer a limited perspective on the qualities of rationales. For example, does the 0.77 \esnli sufficiency reported in \figref{fig:overall_sufficiency} correspond with a similar drop in accuracy, or do the rationales render the model less confident but equally accurate? And how can we distinguish between a highly concise rationale and one bloated with unnecessary information?
We propose extensions of the basic fidelity framework to address these more nuanced questions.

\subsection{Accuracy Evaluation with Rationales}

Existing fidelity metrics measure differences in output probability rather than model performance, prompting the question of what is the practical effect of rationale fidelity. Moreover, they generally involve a model trained on complete texts but then evaluated on reduced texts based on rationales, rendering it unclear what outcome differences we can attribute to the missing information, and what to domain transfer between full and reduced text.

To answer these questions, we compare the accuracy of 
three variant training/evaluation regimes:
1) trained and evaluated on full text (\norationale); 2) trained on full text and evaluated on rationale-only text (\evalrationale); and 3) trained and evaluated on rationale-only text (\trainevalrationale).  

The first variant is standard \roberta model
training and evaluation. The second variant is the typical rationale evaluation setting: trained on full data and evaluated on reduced data.
The third variant seeks to assess what performance gains can arise from model adaptation to the reduced data distribution. Table \ref{tab:rationale_use} summarizes the variants.%
\footnote{We only have human rationales on 1,049 instances in \wikiattack, so we use a different train/dev/test split from \secref{sec:evaluation}.}

\begin{figure}[] 
\centering
  \includegraphics[width=1.0\linewidth]{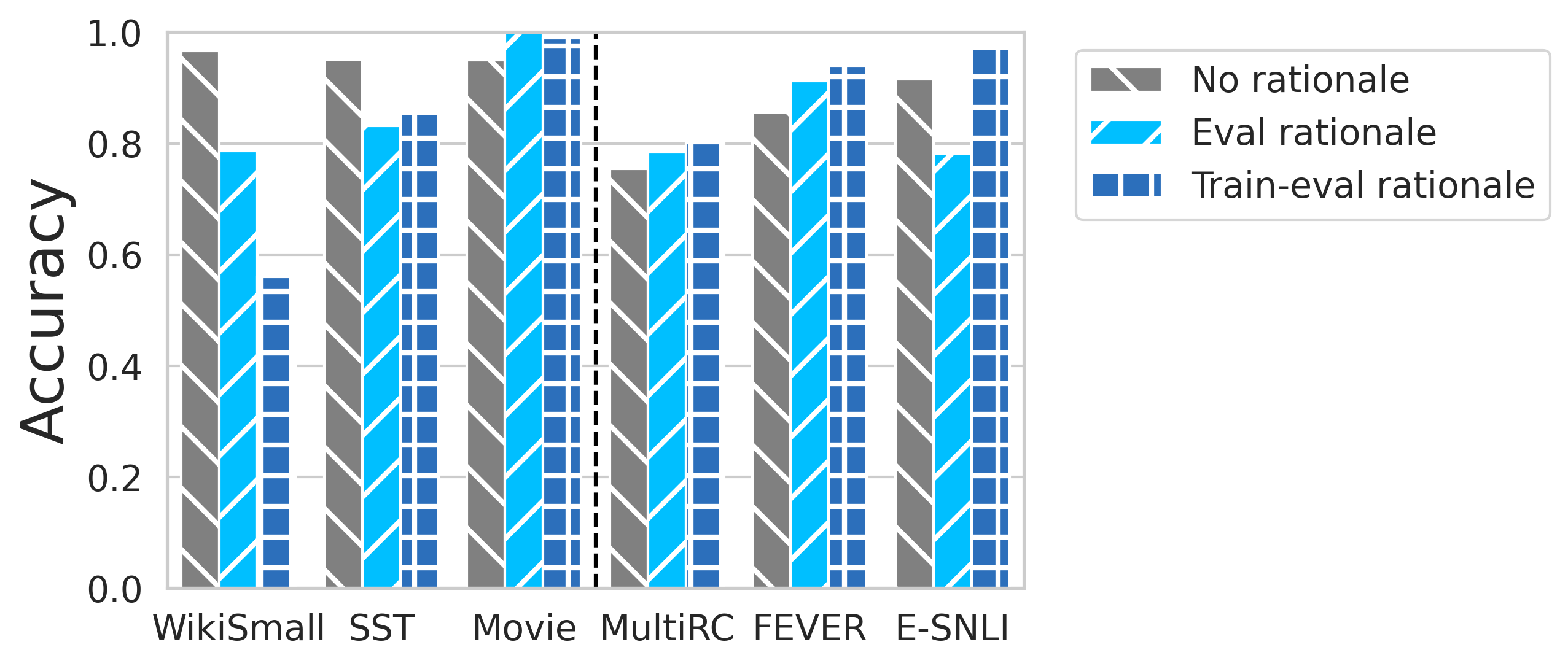}

  \caption{\roberta accuracy depending on whether we adapt models to rationale-only data. Human rationales are effective in improving accuracy in \movie and \esnli, but not in \wikiattack and \sst.
  }
  \label{fig:oracle_suff_bert}
\end{figure}
\begin{table}[t]
    \centering
    \footnotesize
    \begin{tabular}{@{}lp{0.2\linewidth}p{0.2\linewidth}}
    \toprule
    & Training & Testing\\
    \midrule
    No-rationale & No & No \\ 
    Eval-rationale & No & Yes \\
    Train-eval-rationale & Yes & Yes \\
    \bottomrule
    \end{tabular}
    \caption{Use of rationales in different accuracy evaluations. The full-text model uses {\em no} rationale in training.}
    \label{tab:rationale_use}
\end{table}

Comparing the performance of these three models pits the benefits of data completeness (training on full information) against those of in-domain training (training on the same distribution as the evaluation data). If the former proves more valuable we would expect \evalrationale to outperform \trainevalrationale, and vice versa. In either case, we expect \norationale to have the best performance as it benefits from both qualities. 

\figref{fig:oracle_suff_bert} shows some surprising divergences from these expectations.
In four out of six cases, either \trainevalrationale accuracy or \evalrationale accuracy outperforms \norationale accuracy. 

The effect of rationales in evaluation gives yet another perspective on the basic fidelity results presented in \figref{fig:overall_sufficiency}. While the 0.77 sufficiency 
for \esnli corresponds with a significant accuracy drop between \norationale and \evalrationale, the 0.85 sufficiency 
for \multirc corresponds with an \textit{increase} in accuracy across these variants. The almost identical sufficiency of \sst corresponds with a drop. 
``Insufficient'' explanations can improve model performance, which suggests caution in using fidelity based on output probability as the sole arbiter of explanation quality.

The effect of model adaptation has interesting implications as well. We observe an improvement in performance from \evalrationale to \trainevalrationale in 4 out of 6 datasets, significant in the case of \esnli. In 3 out of 4 of these cases, the performance of \trainevalrationale also exceeds that of the 
\norationale setting.

This result is a hopeful sign for the topic area of \textit{learning-from-explanation}, which seeks to use explanations as additional training supervision for models \citep{hancock2018training,zaidan2008modeling}. It tells us that for a majority of our datasets, a perfectly human-mimicking rationale layer could boost the accuracy of a model's predictions. It is even possible that a version of this analysis could be used as a preliminary assessment of the usefulness of a rationale dataset as accuracy-boosting signal, though we leave this for future work.

In summary, 
from a model accuracy perspective, the quality of human rationales is strong for \fever, \multirc, and \movie, mixed for \esnli, and poor for \sst and \wikiattack. 
This provides a somewhat different view from \figref{fig:normalized_bert_sufficiency_by_class}.
For example, human rationales in \multirc has lower (normalized) sufficiency based on output probability than \sst but provide better accuracy sufficiency.

\begin{figure*}[t]
\centering

\begin{subfigure}[t]{0.255\textwidth}
  \centering
    \includegraphics[width=\textwidth]{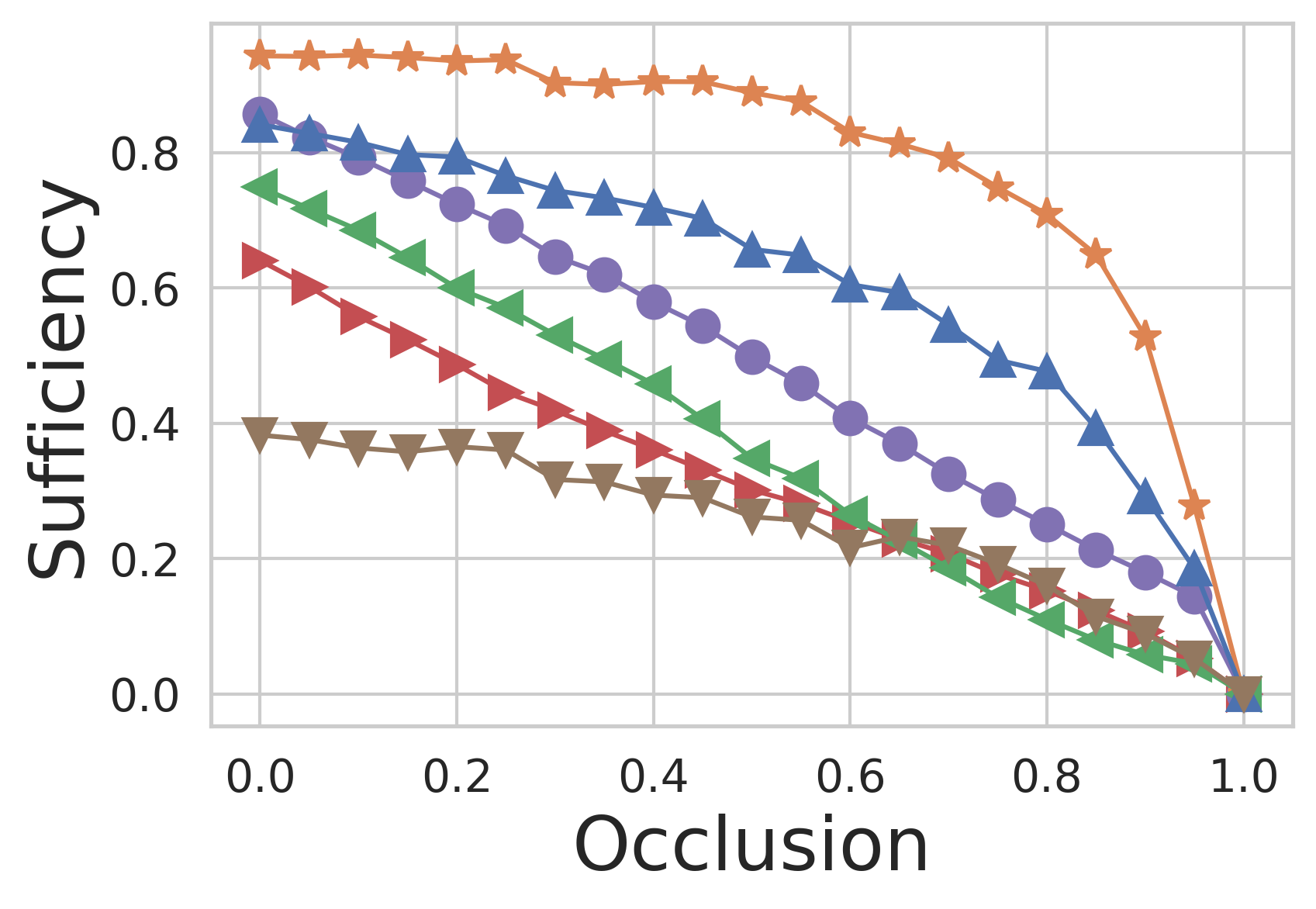} 
  \caption{Sufficiency}
  \label{fig:sufficiency_curves}
\end{subfigure}
\begin{subfigure}[t]{.345\textwidth}
  \centering
    \includegraphics[width=\textwidth]{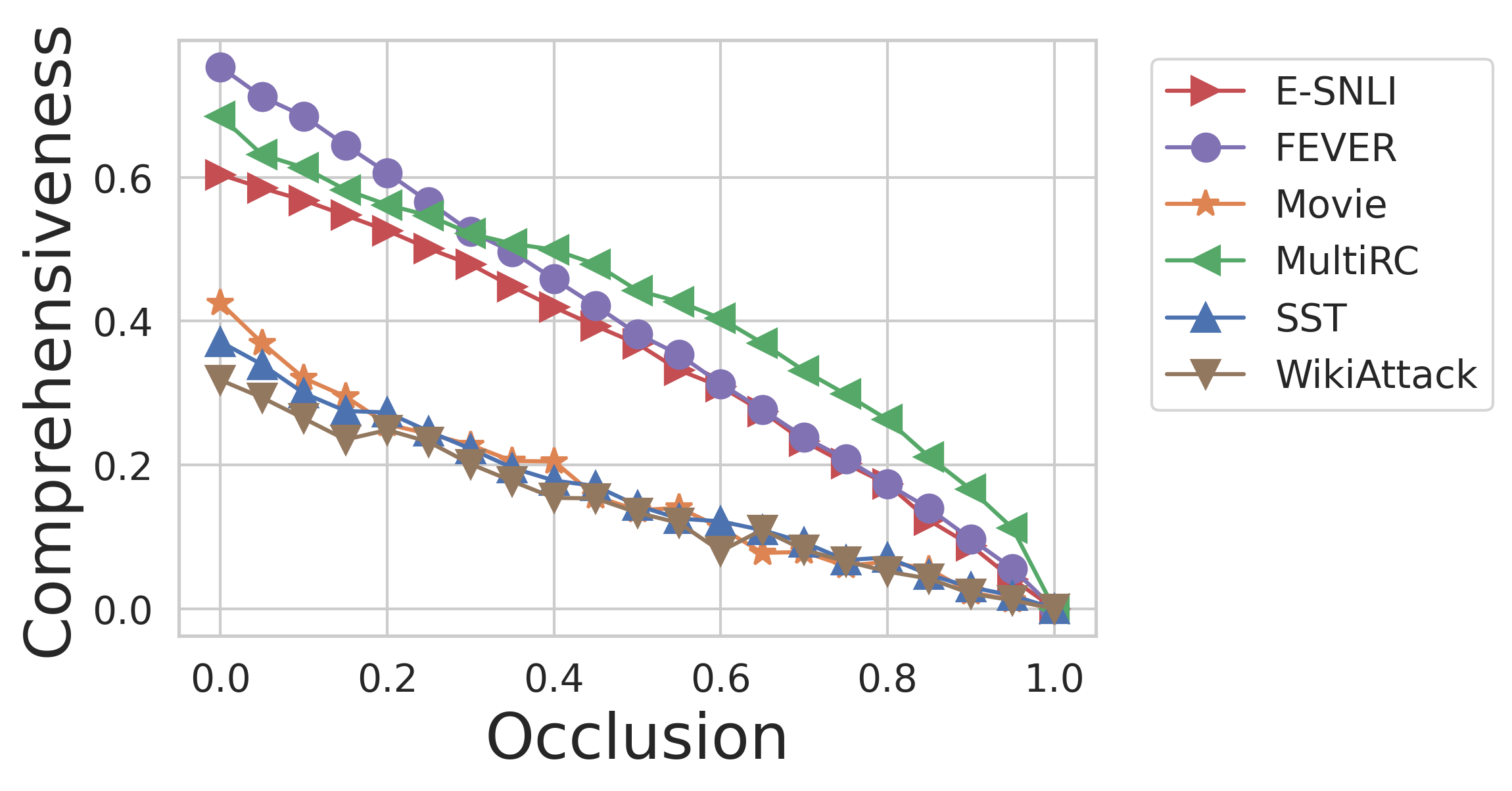}

  \caption{Comprehensiveness}
  \label{fig:comprehensiveness_curves}
\end{subfigure}
\begin{subfigure}[t]{0.38\textwidth}
    \vspace{-0.95in}
    \centering
    \scriptsize
    \begin{tabular}{@{}lp{0.3\linewidth}p{0.3\linewidth}}
    \toprule
    & Sufficiency & Comprehensiveness\\
    \midrule
   WikiAttack & slow drop & fast drop \\ 
    SST & slow drop & fast drop \\
    Movie & slow drop & fast drop \\
   MultiRC & fast drop & fast$\rightarrow$slow drop \\
    FEVER & fast drop & fast drop \\
    E-SNLI & fast drop & slow drop \\
    \bottomrule
    \end{tabular}
    \caption{Summary by dataset.}
    \label{tab:parsimony_results}
\end{subfigure}
\caption{Fidelity curves for all datasets (normalized sufficiency and comprehensiveness).
Human rationales  tend to be redundant in the single-text classification datasets, and dependent in the \qastyle datasets.}
%
%
%
%
%
%
%
%

\label{fig:parsimony_curves}
\end{figure*}

\subsection{Fidelity Curves}
\label{sec:fidelity_curves}

Sufficiency and comprehensiveness 
struggle to convey more fine-grained qualities of human rationales.
One problem that is not revealed by these measures is 
irrelevance. 
A rationale can be crammed with tokens that are not pertinent to prediction and still have high sufficiency and comprehensiveness, the most extreme example being a rationale that comprises the entire text. 

We propose to assess rationale irrelevancy by looking at how sufficiency and comprehensiveness degrade as tokens are removed from the rationale. 
A rationale bloated with many irrelevant tokens should demonstrate a slow dropoff in sufficiency as tokens are removed, since many of these tokens will not be contributory. A rationale with more informational brevity should show a faster drop, as tokens are removed which were needed for prediction. We assess this by creating a ``sufficiency curve'' which traces this degradation at higher and higher occlusion rates.

In general, we suggest that a slow drop in sufficiency can be attributed to irrelevant or redundant tokens,
while a fast drop in sufficiency can be due to dropping tokens that are either individually predictive or pieces of dependencies where  multiple tokens are required to make a prediction. We can tell the difference by looking at the comprehensiveness curve --- if individually predictive tokens are leaked into the rationale complement, the comprehensiveness should fall quickly, while if pieces of dependencies are, it should fall slowly.
\tableref{tab:parsimony_hypothesis} 
summarizes our expectations.

\begin{table}[t]
    \centering
    \footnotesize
    \begin{tabular}{@{}lp{0.3\linewidth}p{0.3\linewidth}}
    \toprule
    & Sufficiency & Comprehensiveness\\
    \midrule
    brevity & fast drop & fast drop \\ 
    redundancy & slow drop & fast drop \\
    irrelevance & slow drop & slow drop \\
    dependency & fast drop & slow drop \\
    \bottomrule
    \end{tabular}
    \caption{Implications of irrelevance and redundancy on sufficiency and comprehensiveness.}
    \label{tab:parsimony_hypothesis}
\end{table}

We construct these fidelity curves as follows: For a given rationale $\rationale$ and each of a series of replacement rates $R={0,0.05,0.1, ... , 1.0}$, we generate a reduced mask $\rationale_r$ 
by randomly setting $r$ fraction of tokens to 0 from the rationale. By calculating the mean normalized sufficiency and comprehensiveness over several trials for each replacement rate, we can draw a ``sufficiency curve'' (\figref{fig:sufficiency_curves}) and a ``comprehensiveness curve'' (\figref{fig:comprehensiveness_curves}).
\movie, \wikiattack, and \sst 
exhibit 
slow drops in their sufficiency curves,
showing that rationales in these datasets contain relatively many irrelevant or redundant tokens, and therefore remain sufficient even as some of their tokens are removed. 
Their comprehensiveness curves complete the story. The curves for all three datasets show relatively fast drops, 
implying redundancy rather than irrelevancy.
In comparison, \esnli, \fever, and \multirc
all display relatively fast drops in sufficiency, implying fewer irrelevant or redundant tokens. 
They demonstrate generally higher comprehensiveness
but somewhat different shapes (\esnli and \multirc mostly show a slow drop, indicating dependence, while \fever shows a fast drop, indicating irrelevance).
The difference here between \fever and \multirc is interesting as they are similar in task, text, and rationale properties (Table \ref{tab:statistics}). A possible explanation is that rationales in \multirc are designed to consist of multiple mutually-dependent sentences whereas those of \fever are single contiguous snippets of the text. 
This greater level of dependency is thus reflected in the slow-dropping comprehensiveness curve of \multirc. 

Hence, we find that human rationales for the three 
classification tasks are characterized by 
redundancy in human rationales, particularly 
\movie.
The three \qastyle datasets, by contrast, are characterized by a relatively high degree of token dependency, explaining their relatively high comprehensiveness in \figref{fig:normalized_bert_comprehensiveness_by_class}.
While this observation is intuitive given the semantics of these tasks, it demonstrates the effectiveness of the proposed fidelity curves.

\section{Related Work}

We summarize additional related work in the following three areas.

\para{Feature attribution.} Feature attribution seeks to explain model behavior by attributing model predictions to specific inputs. Popular techniques include LIME \citep{ribeiro_why_2016}, integrated gradients \citep{sundararajan_axiomatic_2017}, SHAP \citep{lundberg2017unified},
and attention mechanisms \citep{lei2016rationalizing,paranjape_information_2020}.
\para{Human rationales.}
Many recent datasets in NLP have been released with rationales accompanying the document-level labels. 
ERASER \citep{deyoung2019eraser} includes three additional datasets: CoS-E \citep{rajani_explain_2019}, BoolQ \citep{clark_boolq_2019}, and Evidence Inference \citep{lehman_inferring_2019}. Other rationale datasets 
include that of \citet{kaushik_learning_2019} and \citet{sen_human_2020}.

\para{Attribution evaluation.}
A growing amount of work seeks to evaluate the quality of feature attribution. 
Beyond collecting human rationales as a gold-standard, 
a common human-based method is to test the utility of attribution masks in task-based human subject experiments %
\citep{carton_attention-based_2020,lai_human_2019,poursabzi-sangdeh_manipulating_2018,lage_evaluation_2018,lai+liu+tan:20}. %

Automatic model-based metrics
beyond
sufficiency and comprehensiveness
include local model fidelity \citep{ribeiro_why_2016}, switching point \citep{nguyen_comparing_2018}, and area-over-the-perturbation-curve \citep{samek_evaluating_2017}. 

\section{Concluding Discussion}

Human explanations contain a lot of promise. The explainable AI community hopes to use them as a guide for 
evaluating model explanations
and, possibly, for teaching models to make 
robust and well-reasoned decisions. In this work, we contribute to that effort by analyzing human rationales through the lens of automatic rationale evaluation methods, namely, sufficiency and comprehensiveness.
We find that human rationales do not necessarily have high sufficiency or comprehensiveness.

\para{Interpreting fidelity variance.}
Furthermore, there exists significant variance across datasets and classes.
In \secref{sec:fidelity_curves}, we speculate that some of these differences (e.g., dependency) can be explained by the semantic differences between classification and \qastyle tasks.

However, with such a small sample size of datasets ($n=6$), it is difficult to determine whether these differences are due solely to task type or to other factors such as annotation instructions or individual dataset semantics. \wikiattack and \esnli, for example, display class asymmetry in their rationales, which likely contribute to their outlier status in \figref{fig:normalized_bert}  and \ref{fig:oracle_suff_bert} respectively. 
As we note in \figref{fig:overall}, modeling outcomes also have a heavy impact on explanation fidelity.  While \esnli comprises an even class balance, our model learns a strong bias in favor of the neutral class, which contributes to a class imbalance in fidelity for that dataset (\figref{fig:normalized_bert}). 

As more human-rationale datasets are released, it will become increasingly possible to categorize them by rationale properties. Our goal 
is 
to highlight the variance in these properties and call for more widespread empirical evaluations thereof. 

\para{Actionable implications.}
When human rationales are found to be unfaithful, this can mean that either they fail to capture relevant signal, or that the model improperly utilizes that signal, perhaps as a result of learning spurious associations. In either case, analysis can expose inconsistencies between human and model understanding of the task. 

We propose three ways to extend fidelity metrics: normalization, model adaptation, and random ablation. Each addresses one shortcoming of the basic metric: normalization addresses the differences in class biases across models, adaptation the problem of domain inconsistency between full and rationale-only data, and ablation the inability of existing metrics to capture qualities like redundancy. While not all of these issues are salient for every application involving rationale fidelity, we offer them as potential solutions where necessary. 

Overall, our results suggest that the idea of one-size-fits-all fidelity benchmarks might be problematic: human rationales may not be simply treated as gold standard.
We need to design careful procedures to collect human rationales, understand properties of the resulting human rationales, and cautiously interpret the evaluation metrics.

\para{Acknowledgments.}
We thank helpful comments from anonymous reviewers.
This work was supported in part by research awards from Amazon and Salesforce, and NSF IIS-1927322, 1941973.

\bibliographystyle{acl_natbib}
\bibliography{emnlp2020}

\appendix

\appendix

\section{Derivation of Rationales for \sst}

The Stanford Sentiment Treebank (SST) consists of 9,620 short movie review snippets formatted as syntactic trees with a sentiment label in [-2,2] for each node, ranging from the single-token leaf nodes to the top-level node corresponding to the whole snippet.  

We use a heuristic algorithm for flattening this representation into a 1-dimensional rationale for each document: beginning with the top node and traversing the tree in a breadth-first manner, we consider a node to be part of the rationale if the magnitude of its sentiment is greater than that of any of its descendants. That is, if the sentiment of a node cannot be explained by any of its syntactic constituents, then we consider it to be explanatory and include it in the top-level rationale.

Practically speaking, this results in a rationale dataset that is comprehensive by design, including all high-sentiment words and phrases that could explain the overall sentiment of each snippet. Table \ref{tab:sst_examples} shows a few examples of the resultant rationales.

\begin{table*}[]
\small
\begin{tabular}{@{}p{.90\textwidth}p{.07\textwidth}@{}}
\toprule
\multicolumn{1}{c}{\textbf{Rationale}}                                                                                                                                                                  & \textbf{Class} \\ \midrule
All the performances are \highlightrationale{top notch} and , once you get through the accents , All or Nothing becomes an emotional , though \highlightrationale{still positive ,} wrench of a sit .   & Pos            \\
While \highlightrationale{surprisingly sincere} , this average little story is adorned with some \highlightrationale{awesome} action photography and surfing                                           & Pos            \\ 
A \highlightrationale{dreary rip-off} of \highlightrationale{Goodfellas} that serves as a \highlightrationale{muddled and offensive cautionary tale} for Hispanic \highlightrationale{Americans}       & Neg            \\
A \highlightrationale{long-winded and stagy} session of romantic contrivances \highlightrationale{that never} really gels like the \highlightrationale{shrewd feminist fairy tale} it could have been  & Neg            \\ \bottomrule
\end{tabular}
\caption{Example SST rationales generated by heuristic flattening procedure.}
\label{tab:sst_examples}
\end{table*}

\section{Model Implementation Details}

We consider the following models:

\begin{itemize}[itemsep=0pt,leftmargin=*,topsep=-2pt]
    \item Logistic regression. 
    We use the scikit-Learn  implementation of logistic regression \citep{pedregosa_scikit-learn_2011}, scanning across regularization constant ($C=\{0.001, 0.01, 0.1, 1, 10, 100, 1000\}$).
  \item Random forest. We use the scikit-Learn implementation of random forests, scanning across number of estimators ($\{16, 32, 64, 128, 256, 512\}$). 
  \item LSTM \citep{hochreiter1997long}. We use the Pytorch \citep{paszke_automatic_2017} implementation of a 1-layer BiLSTM, tuning across hidden layer size ($\{100,200,300\}$) and learning rate ($\{5e^{-4},1e^{-3},2e^{-3}\}$).
  \item RoBERTA \citep{liu_roberta:_2019}. We use the HuggingFace \citep{wolf_huggingfaces_2020} pretrained distribution of this model with roughly 117m parameters. We tune the learning rate across values $\{5e^{-6},1e^{-5},2e^{-5}\}$, with 50 linear warmup steps.
\end{itemize}

We train all LSTM models for 10 epochs and RoBERTa models for 5 epochs, tuning on development set accuracy. All neural network training was done on two 24G Nvidia Titan RTX GPUs. Training time varied from dataset to dataset, from minutes for \sst to roughly 6 hours per model for \esnli.

To apply masking, 
we simply remove the tokens corresponding with 0s in the rationale mask.
We always keep special tokens such as [CLS] and [SEP].

Following \citet{deyoung2019eraser}, we flatten the three \qastyle datasets to single documents by simply appending the query to the document with a ``[SEP]'' token.

\section{Eraser Sufficiency/Comprehensiveness vs. Our Definitions}

\begin{figure*}[]
\centering
\begin{subfigure}[t]{.30\textwidth}
  \centering
    \includegraphics[width=\linewidth]{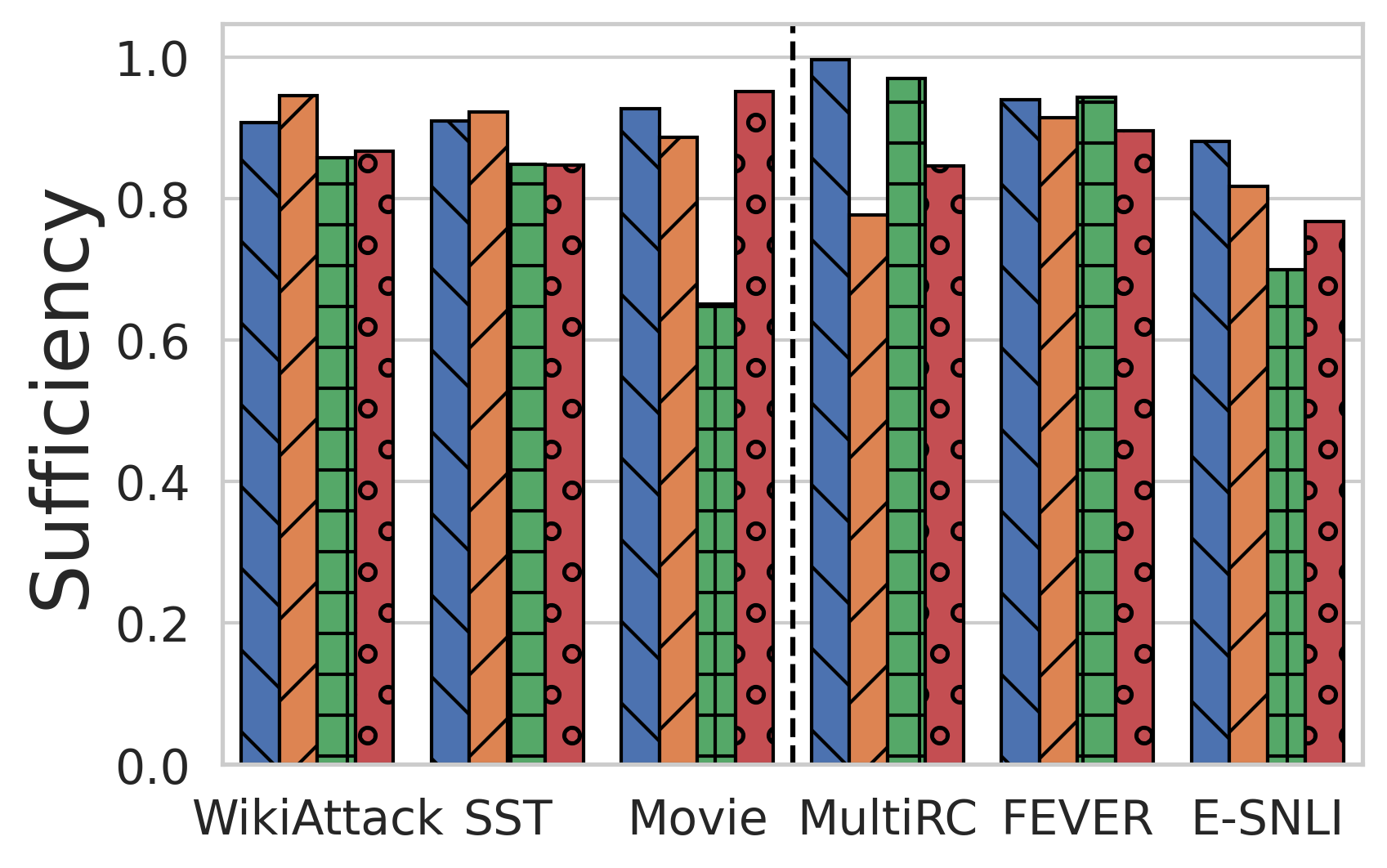}

  \caption{Clipped sufficiency}
  \label{fig:clipped_sufficiency}
\end{subfigure}
\begin{subfigure}[t]{.30\textwidth}
  \centering
      \includegraphics[width=\linewidth]{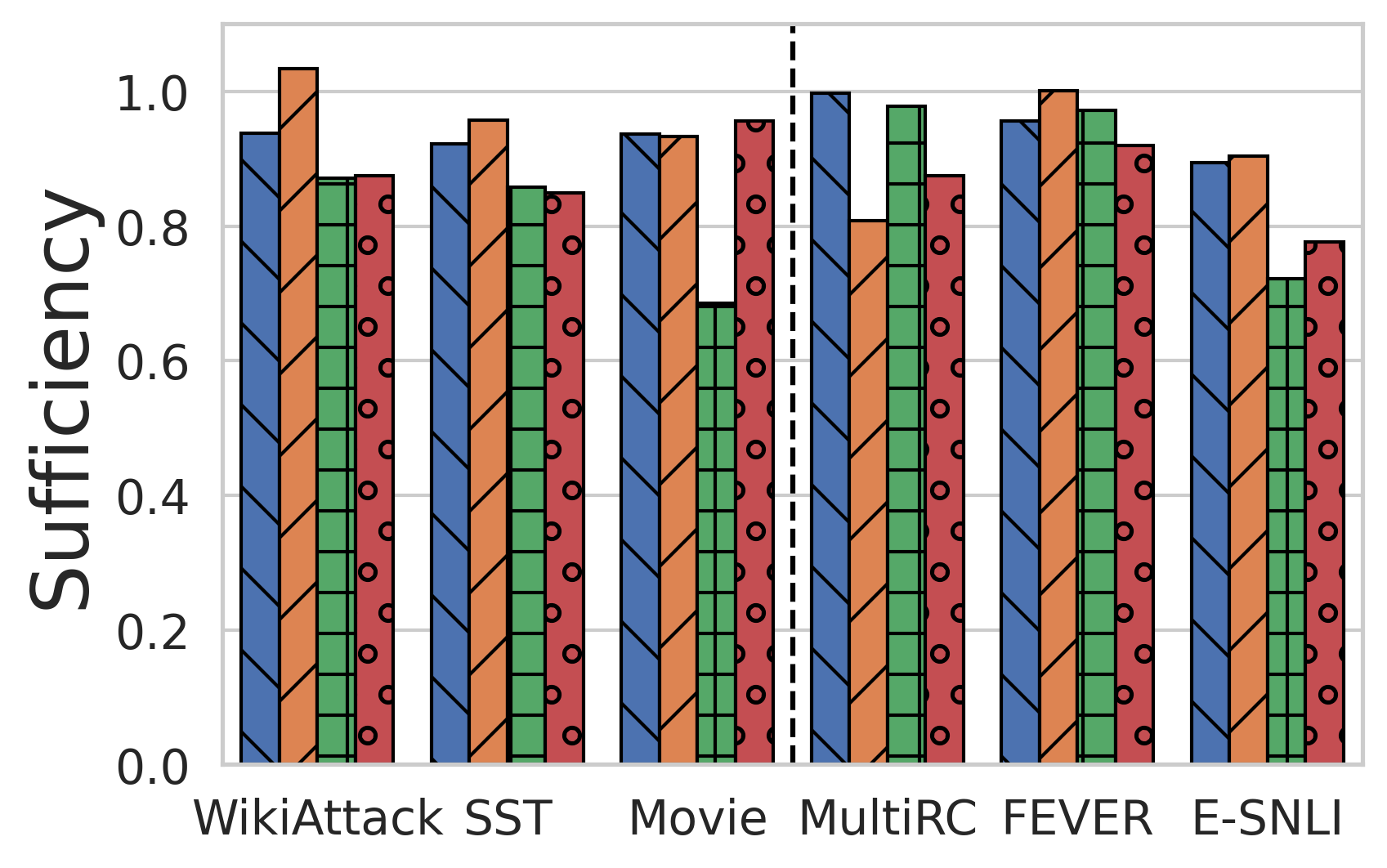}

  \caption{ERASER sufficiency}
  \label{fig:eraser_sufficiency}
\end{subfigure}

\caption{Clipped sufficiency vs. ERASER sufficiency.
}
\label{fig:clipped_versus_eraser_sufficiency}
\end{figure*}

\begin{figure*}[t]
\centering
\begin{subfigure}[t]{0.30\textwidth}
  \centering
    \includegraphics[width=\linewidth]{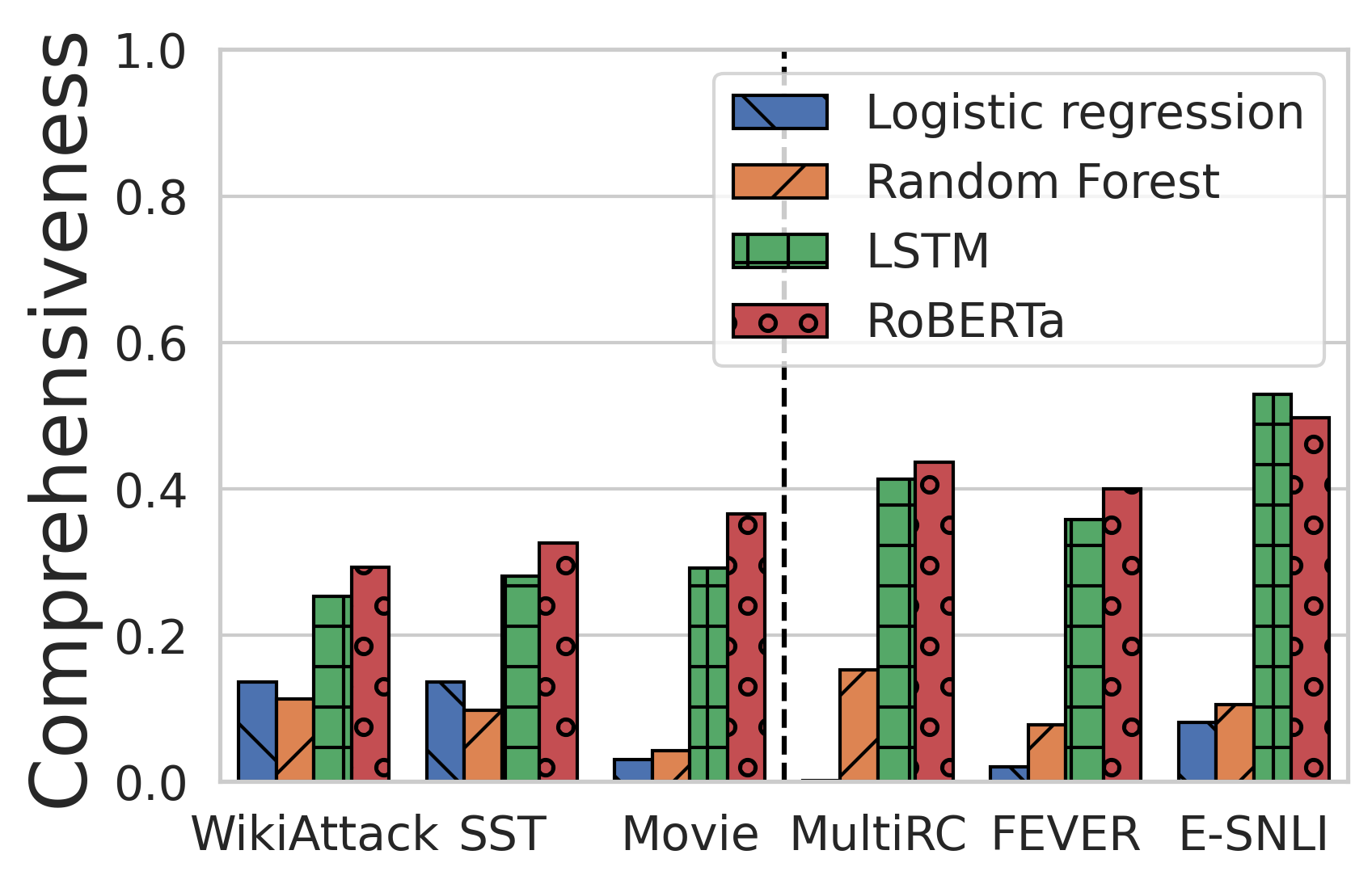}

  \caption{Clipped comprehensiveness}
  \label{fig:clipped_comprehensiveness}
\end{subfigure}
\begin{subfigure}[t]{0.32\textwidth}
  \centering
    \includegraphics[width=\linewidth]{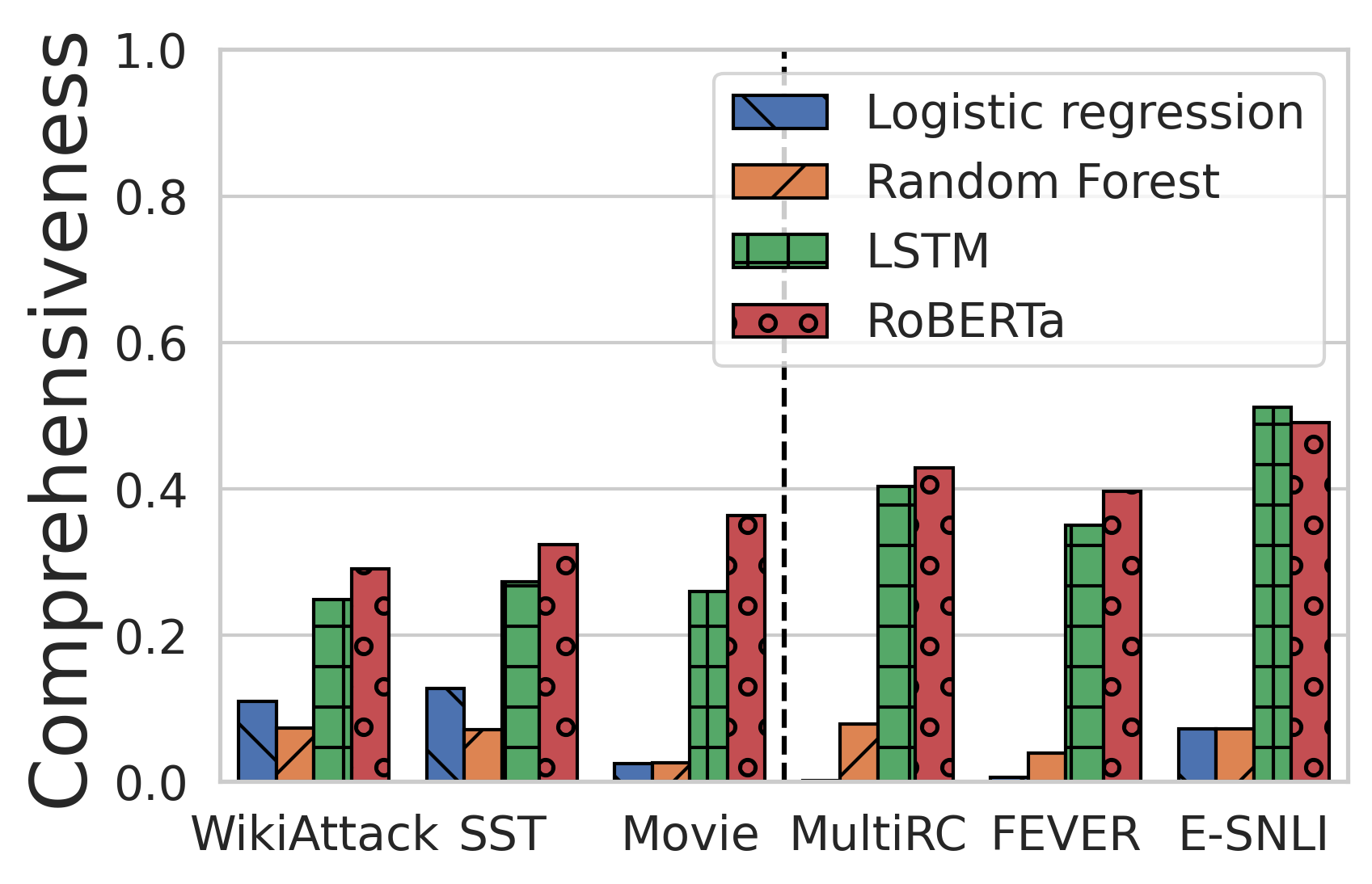}

  \caption{ERASER comprehensiveness}
  \label{fig:eraser_comprehensiveness}
\end{subfigure}
\caption{Clipped comprehensiveness vs. ERASER comprehensiveness.
}
\label{fig:clipped_versus_eraser_comprehensiveness}
\end{figure*}

Our definition of sufficiency and comprehensiveness diverge from that of \citet{deyoung2019eraser} in clipping the absolute difference between the full and rationalized class probability. This choice erases negative probability differences, cases where the rationalization makes the predicted class more probable than it already was. We do this as a way to bound fidelity metrics between 0 and 1.
It also serves to simplify the mathematics of normalization, but practically speaking we find that it makes little difference (\figref{fig:clipped_versus_eraser_sufficiency} and \figref{fig:clipped_versus_eraser_comprehensiveness}). 

\begin{figure*}[]
\centering
\begin{subfigure}[]{.38\textwidth}
  \centering
  \includegraphics[width=0.9\linewidth]{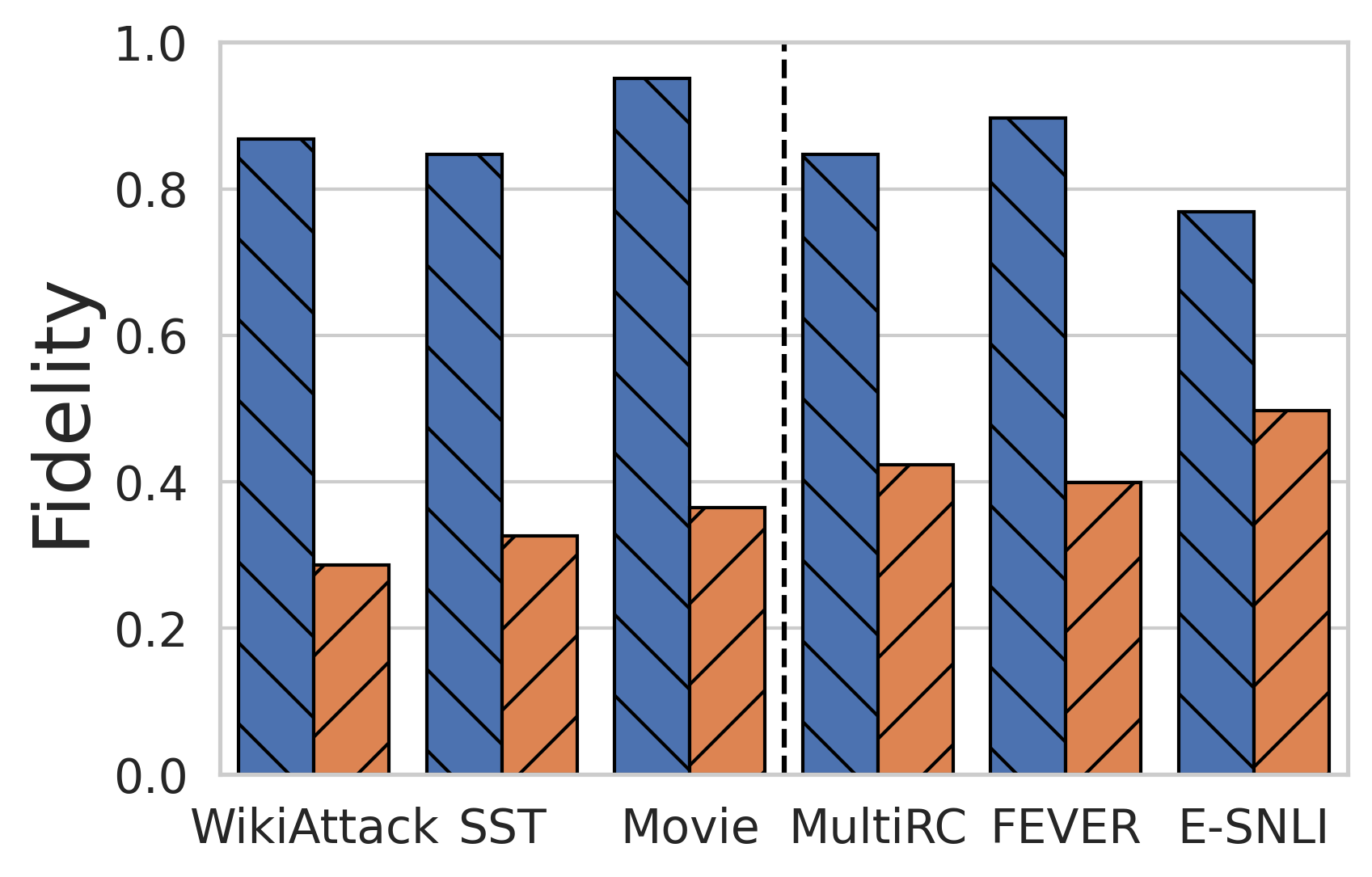}

  \caption{Non-normalized fidelity}
  \label{fig:non_normalized_fidelity}
\end{subfigure}
\begin{subfigure}[]{0.58\textwidth}
  \centering
  \includegraphics[width=0.9\linewidth]{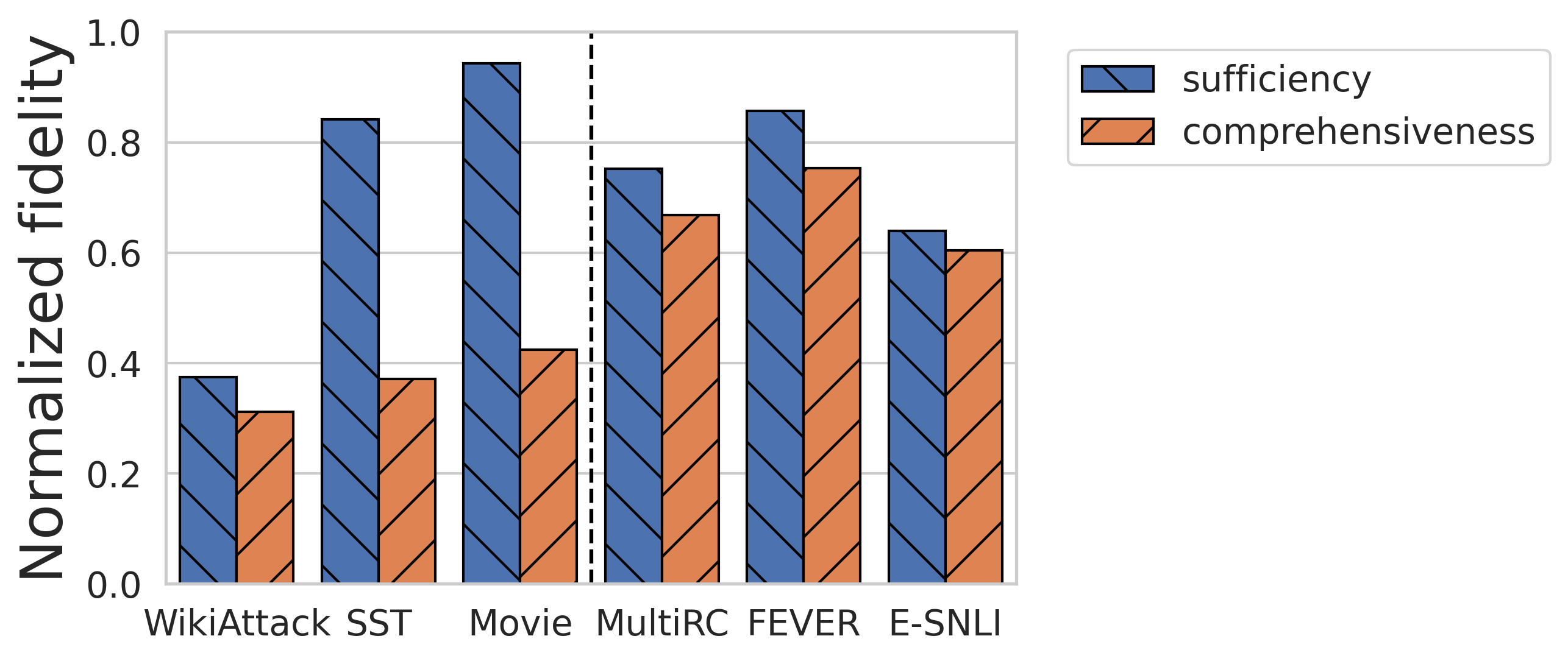}
  \caption{Normalized fidelity}
  \label{fig:normalized_fidelity}
\end{subfigure}

\caption{Non-normalized versus normalized fidelity.}
\label{fig:normalized_versus_nonnormalized_fidelity}
\end{figure*}

\section{The Effect of Normalization}

We discuss the effect of normalization by class in \secref{sec:metrics}. \figref{fig:normalized_versus_nonnormalized_fidelity} compares the non-normalized against the normalized fidelity at the dataset level. This view makes clear the comprehensiveness gap between the classification datasets and the \qastyle datasets, and shows a wider range of  sufficiency scores among the six datasets, when accounting for model bias. 

\begin{figure*}[t]
\centering
\begin{subfigure}[t]{0.30\textwidth}
  \centering
    \includegraphics[width=\linewidth]{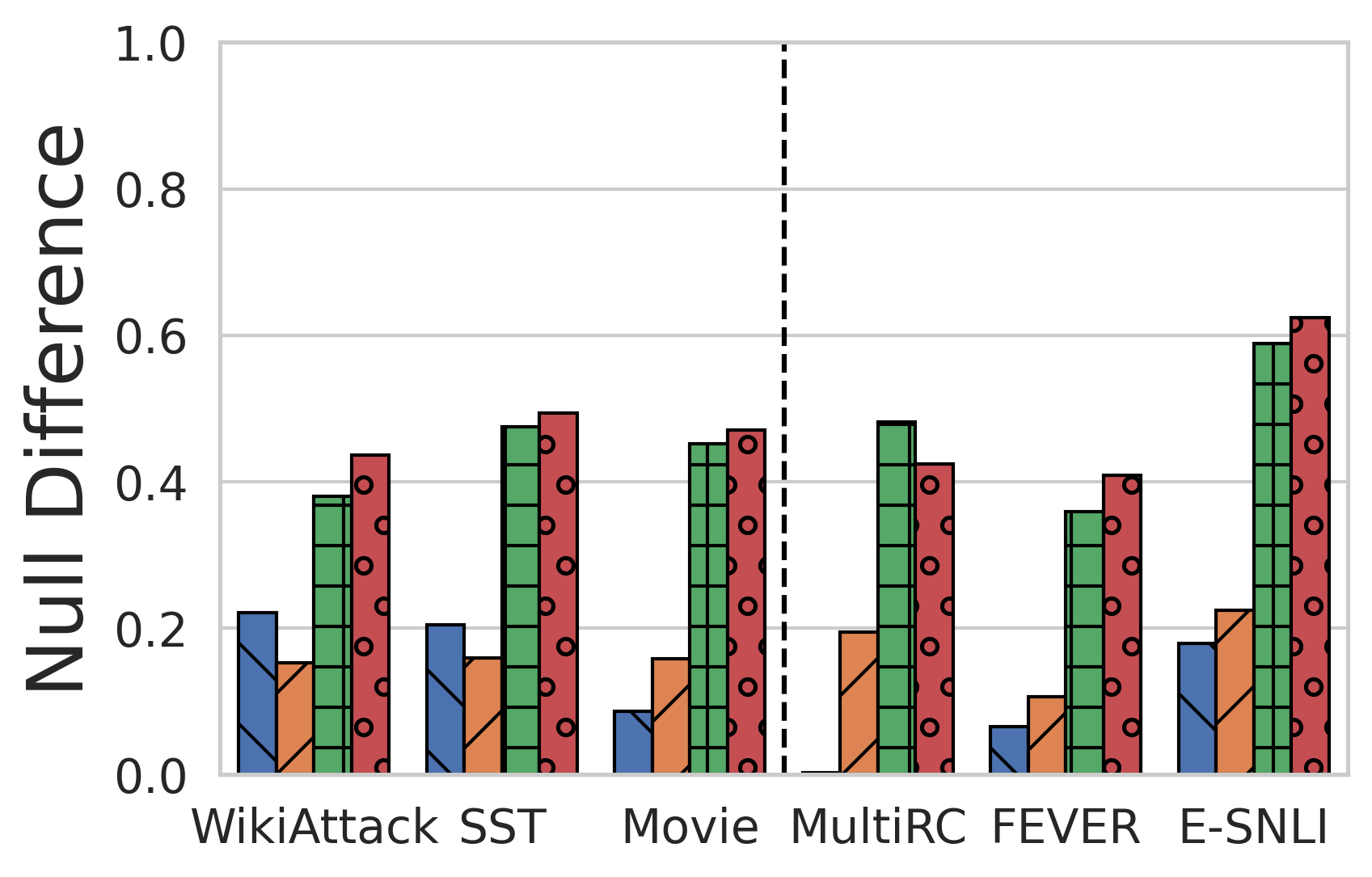}

  \caption{Null difference}
  \label{fig:null_difference_all_models}
\end{subfigure}
\begin{subfigure}[t]{0.30\textwidth}
  \centering
    \includegraphics[width=\linewidth]{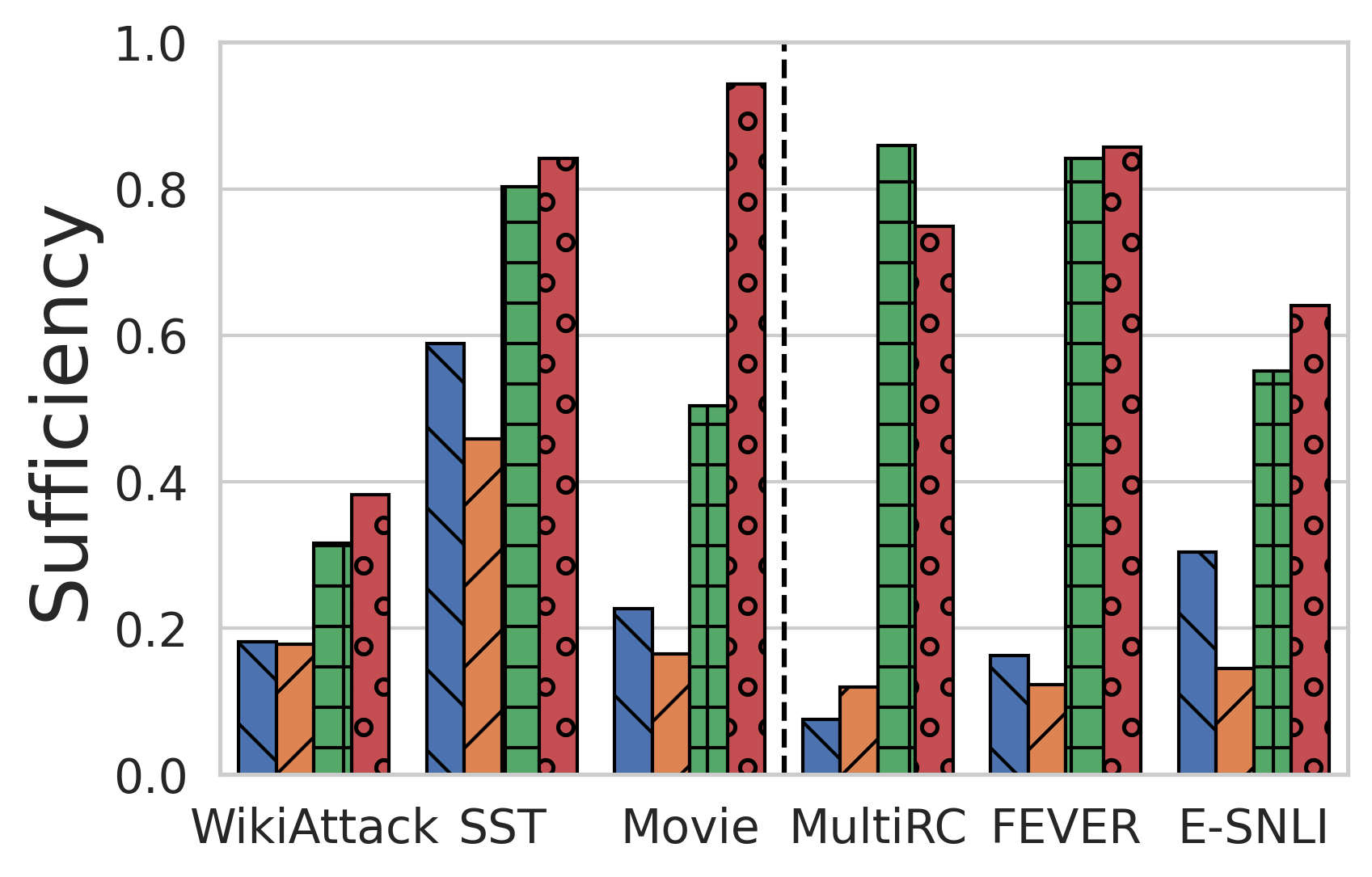}

  \caption{Normalized sufficiency}
  \label{fig:normalized_sufficiency_all_models}
\end{subfigure}
\begin{subfigure}[t]{0.30\textwidth}
  \centering
    \includegraphics[width=\linewidth]{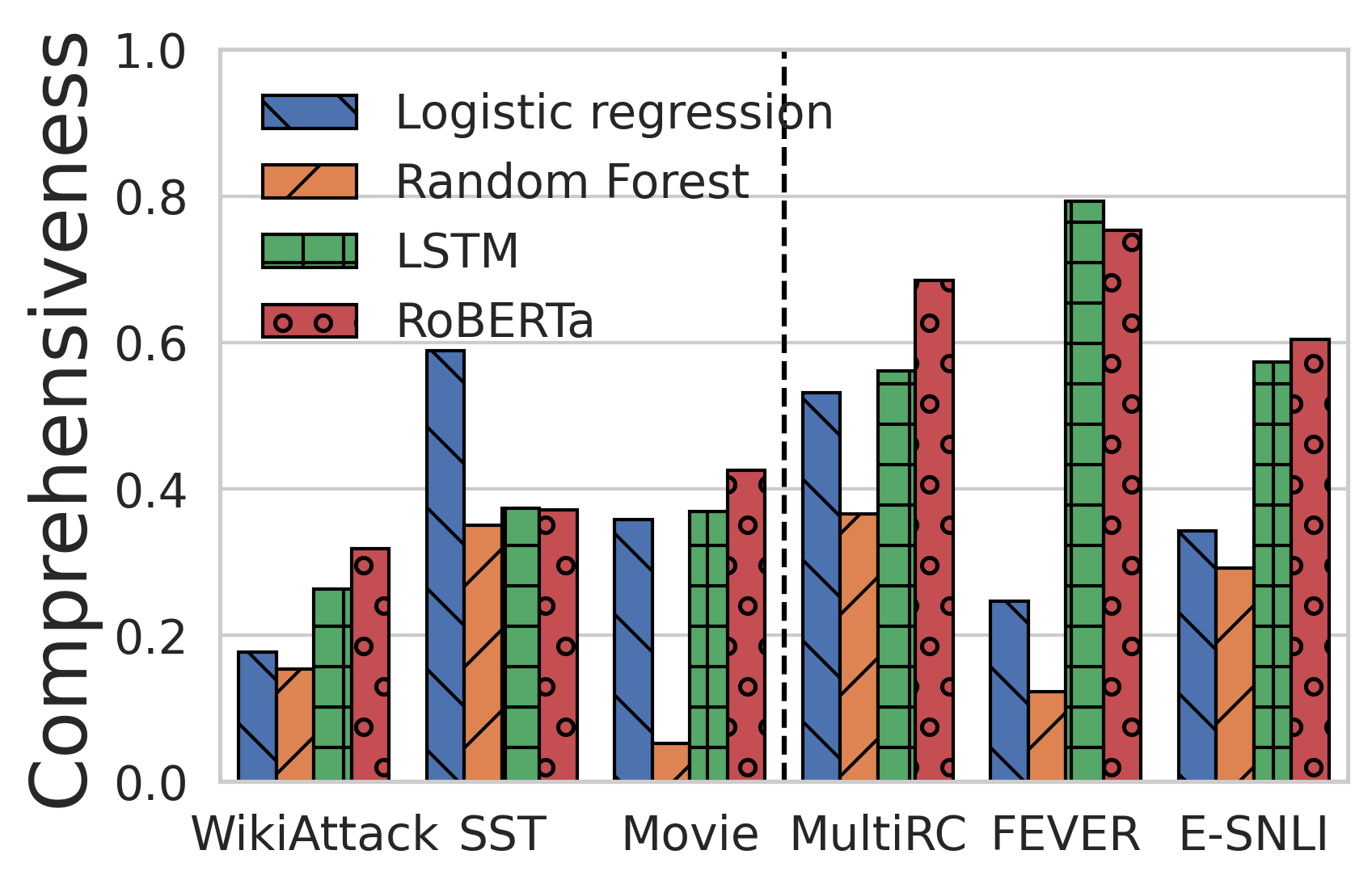}

  \caption{Normalized comprehensiveness}
  \label{fig:normalized_comprehensiveness_all_models}
\end{subfigure}
\caption{Normalized fidelity for all models.}
\label{fig:normalized_fidelity_all_models}
\end{figure*}

\figref{fig:normalized_fidelity_all_models} shows the effect of normalization on fidelity scores for all models. We can see that it corrects the trend of weaker models showing better sufficiency that we observe in \figref{fig:overall_sufficiency}, though logistic regression shows very high sufficiency and comprehensiveness for \sst. Upon investigation, we find that this is because this model tends to have low confidence, often producing class probabilities between 0.5 and 0.7. This situation leads to relatively small null differences (\figref{fig:null_difference_all_models}),
which leads to the high observed comprehensiveness.
In comparison, the null difference is substantially greater in deep models.

\section{The Effect of Hyperparameters and Training}

We largely focus on \roberta in this study because it is close the current state-of-the-art for NLP. However, we do some additional analysis on the other three models. 

\begin{figure*}[]
\centering
\begin{subfigure}[t]{0.28\textwidth}
  \centering
\includegraphics[ width=\textwidth]{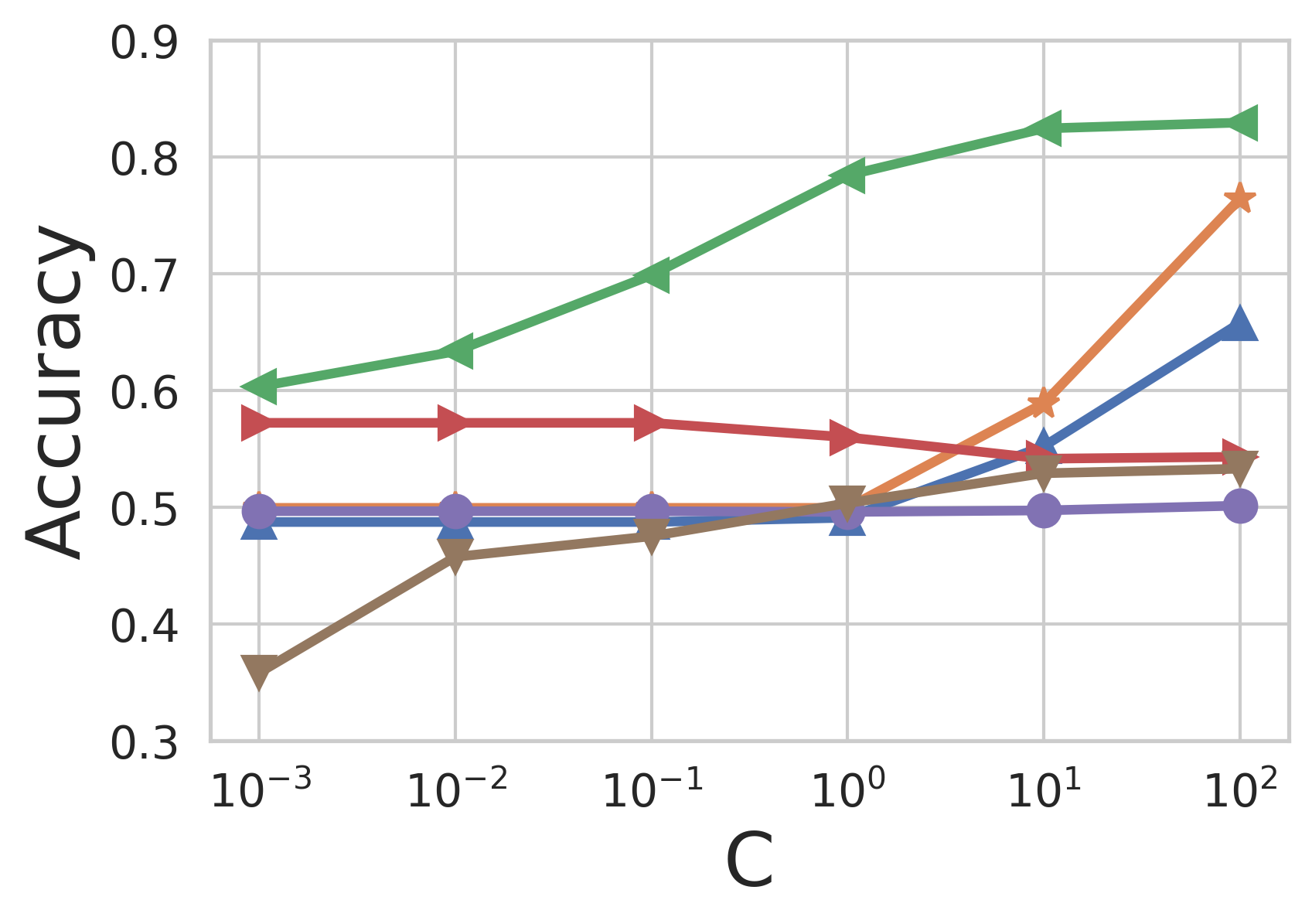} 
  \caption{Accuracy}
  \label{fig:lr_hp_accuracy}
\end{subfigure}
\begin{subfigure}[t]{.28\textwidth}
  \centering
\includegraphics[ width=\textwidth]{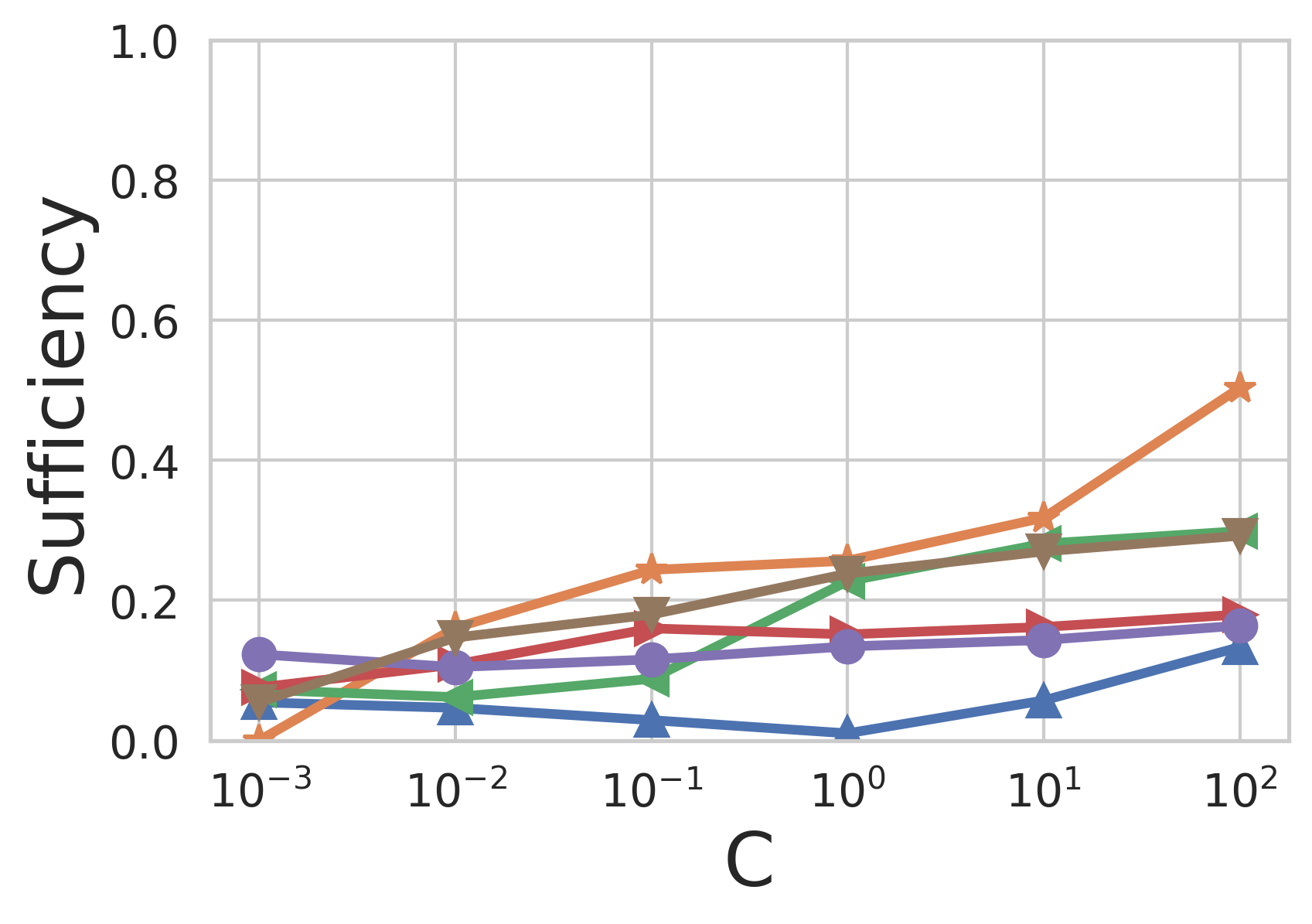} 

  \caption{Normalized sufficiency}
  \label{fig:lr_hp_sufficiency}
  \vspace{0.7em}
\end{subfigure}
\begin{subfigure}[t]{0.38\textwidth}
  \centering
\includegraphics[ width=\textwidth]{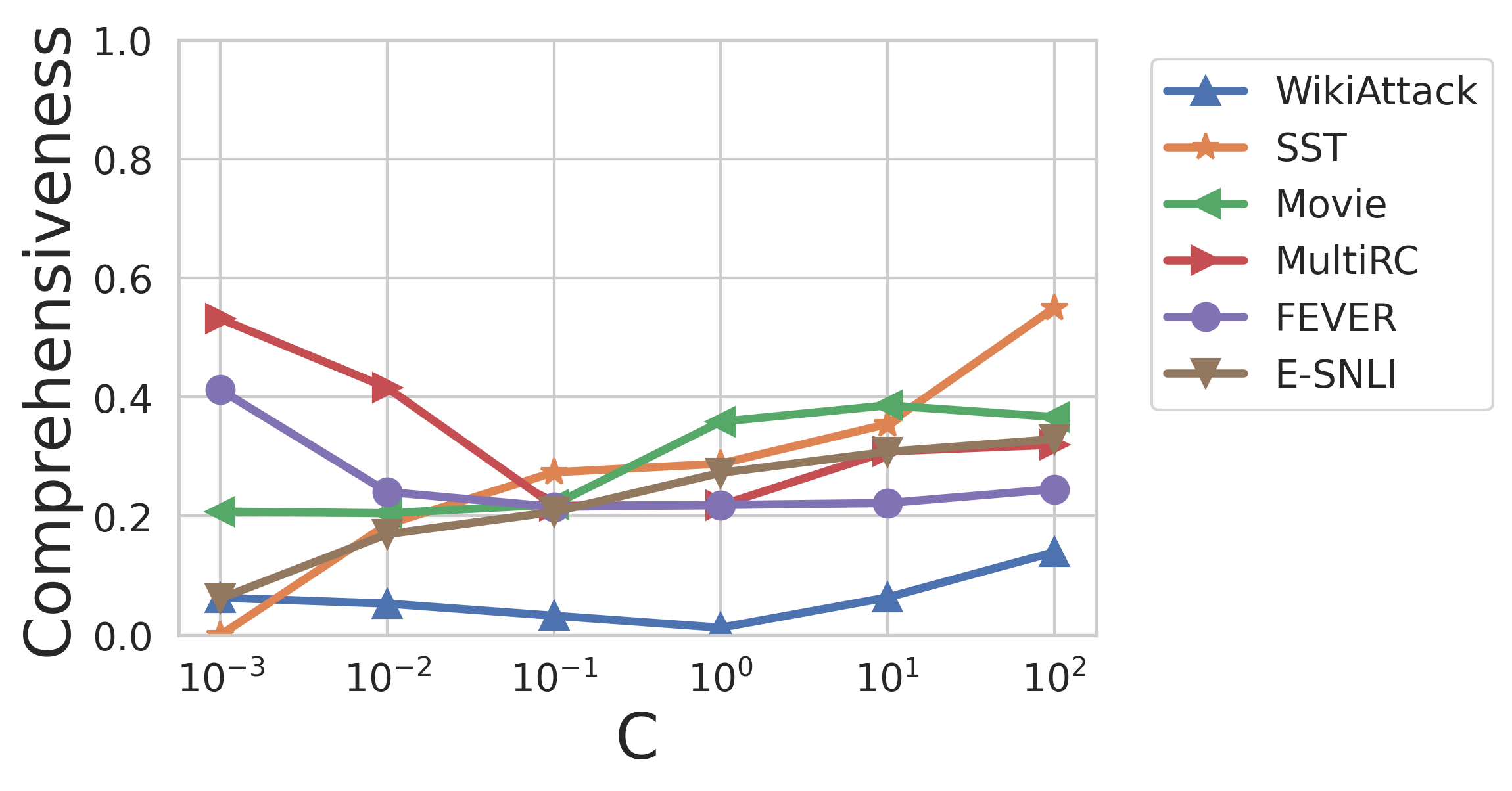} 

  \caption{Normalized comprehensiveness}
  \label{fig:lr_hp_comprehensiveness}
\end{subfigure}

\caption{Accuracy, sufficiency and comprehensiveness of logistic regressions models by regularization term \textit{C}.}
\label{fig:lr_fidelity_by_c}
\end{figure*}

 \figref{fig:lr_fidelity_by_c} shows how accuracy and rationale fidelity change with the value of the $C$ regularization hyperparameter for the logistic regression model. Both the normalized sufficiency and comprehensiveness rise with model accuracy. The outlier is \multirc, which is unable to achieve nontrivial accuracy, but which nevertheless experiences a rise in  rationale fidelity. 

\begin{figure*}[]
\centering
\begin{subfigure}[t]{0.28\textwidth}
  \centering
    \includegraphics[ width=\textwidth]{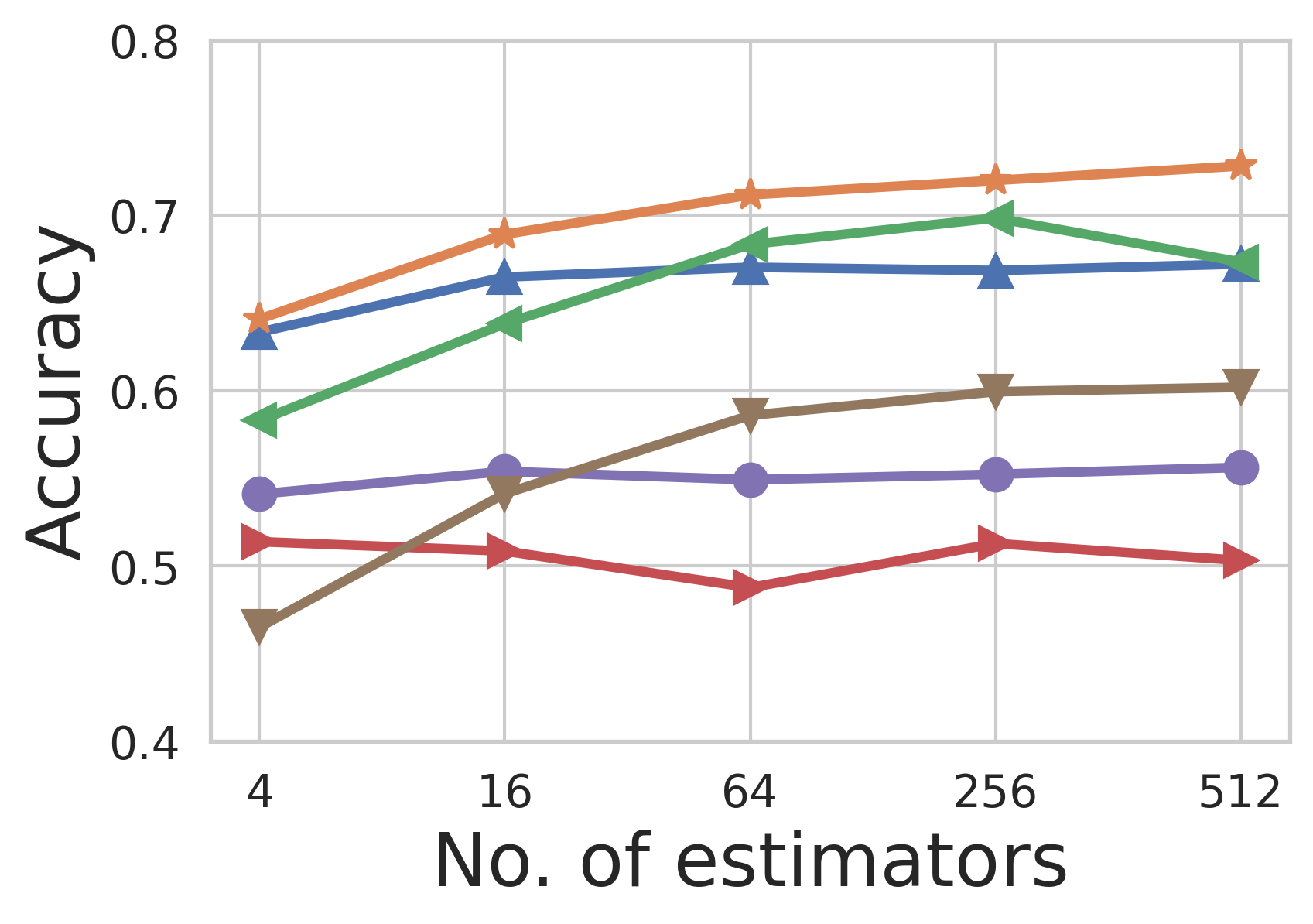} 

  \caption{Accuracy}
  \label{fig:rf_hp_accuracy}
\end{subfigure}
\begin{subfigure}[t]{.28\textwidth}
  \centering
\includegraphics[width=\textwidth]{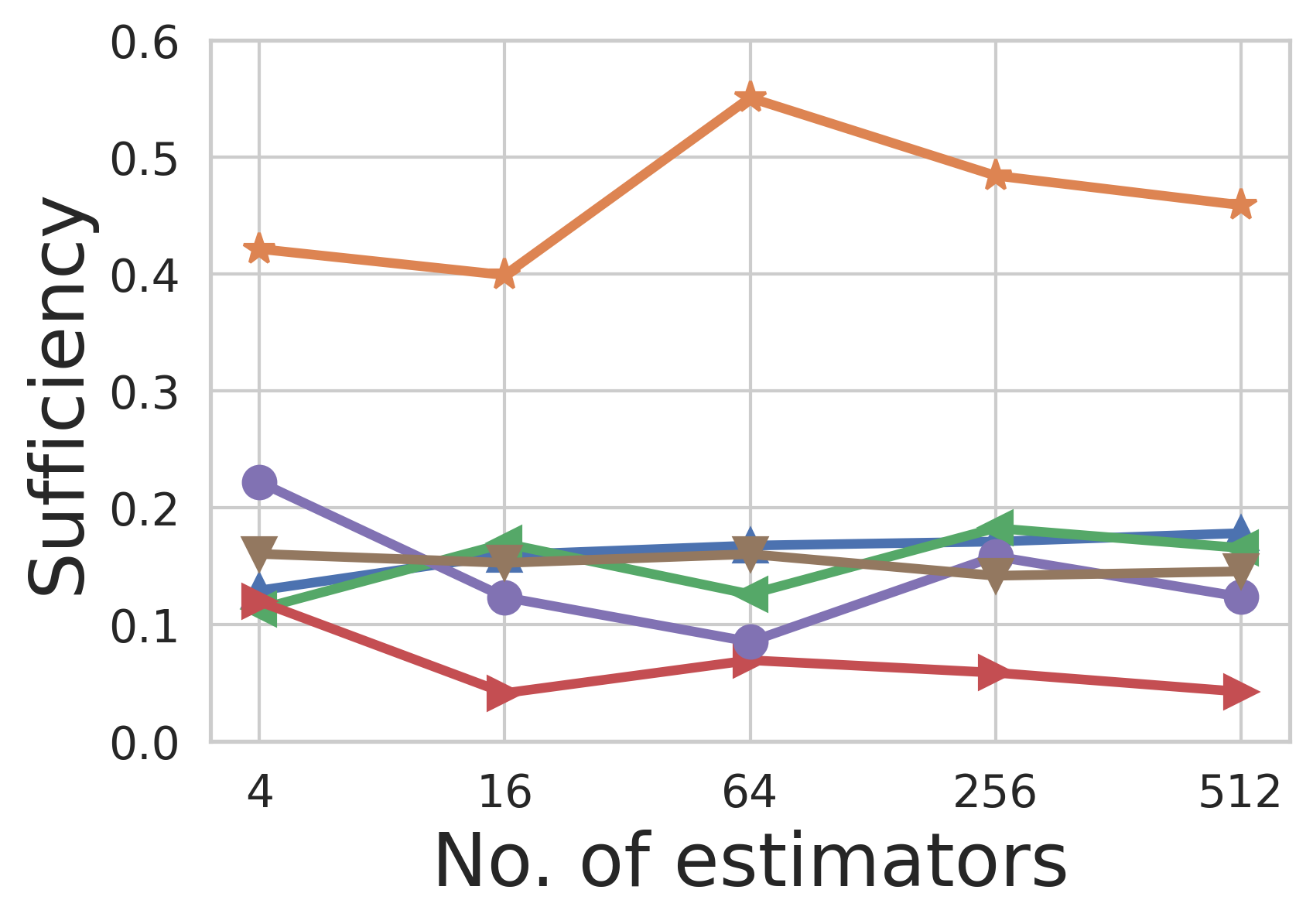} 
  \caption{Normalized sufficiency}
  \label{fig:rf_hp_sufficiency}
  \vspace{0.7em}
\end{subfigure}
\begin{subfigure}[t]{0.38\textwidth}
  \centering
\includegraphics[width=\textwidth]{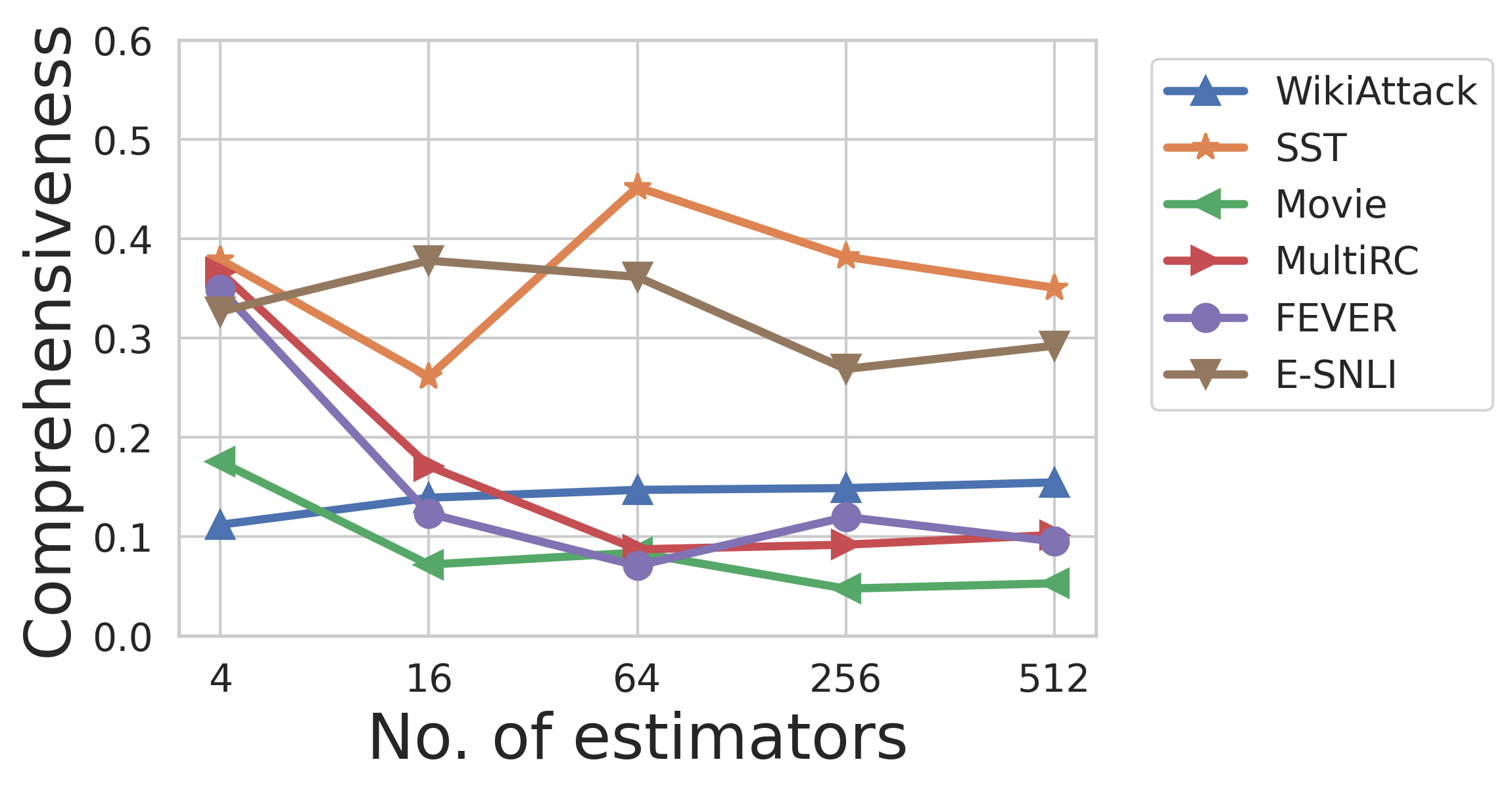} 
  \caption{Normalized comprehensiveness}
  \label{fig:rf_hp_comprehensiveness}
\end{subfigure}

\caption{Accuracy, sufficiency and comprehensiveness of random forest models by number of estimators.}
\label{fig:rf_fidelity_by_n_estimators}
\end{figure*}

The trends are less clear in \figref{fig:rf_fidelity_by_n_estimators}, which tracks accuracy and fidelity over a range of numbers of estimators for the model. This may be because the accuracy of these models does not improve much with the increase in estimators.

\begin{figure*}[]
\centering
\begin{subfigure}[]{0.28\textwidth}
  \centering
    \includegraphics[width=\textwidth]{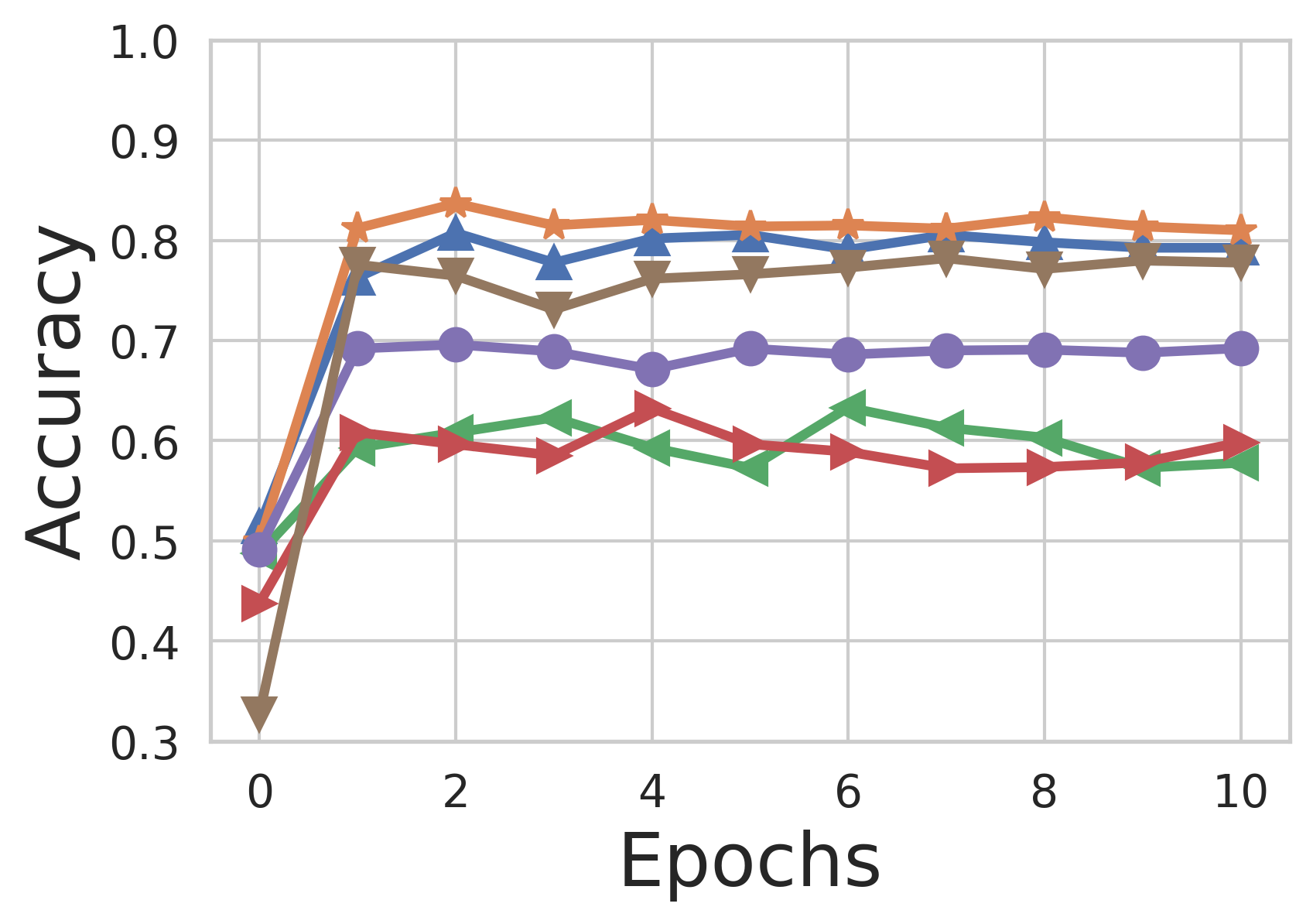} 

  \caption{Accuracy}
  \label{fig:lstm_epoch_accuracy}
\end{subfigure}
\begin{subfigure}[]{.28\textwidth}
  \centering
    \includegraphics[width=\textwidth]{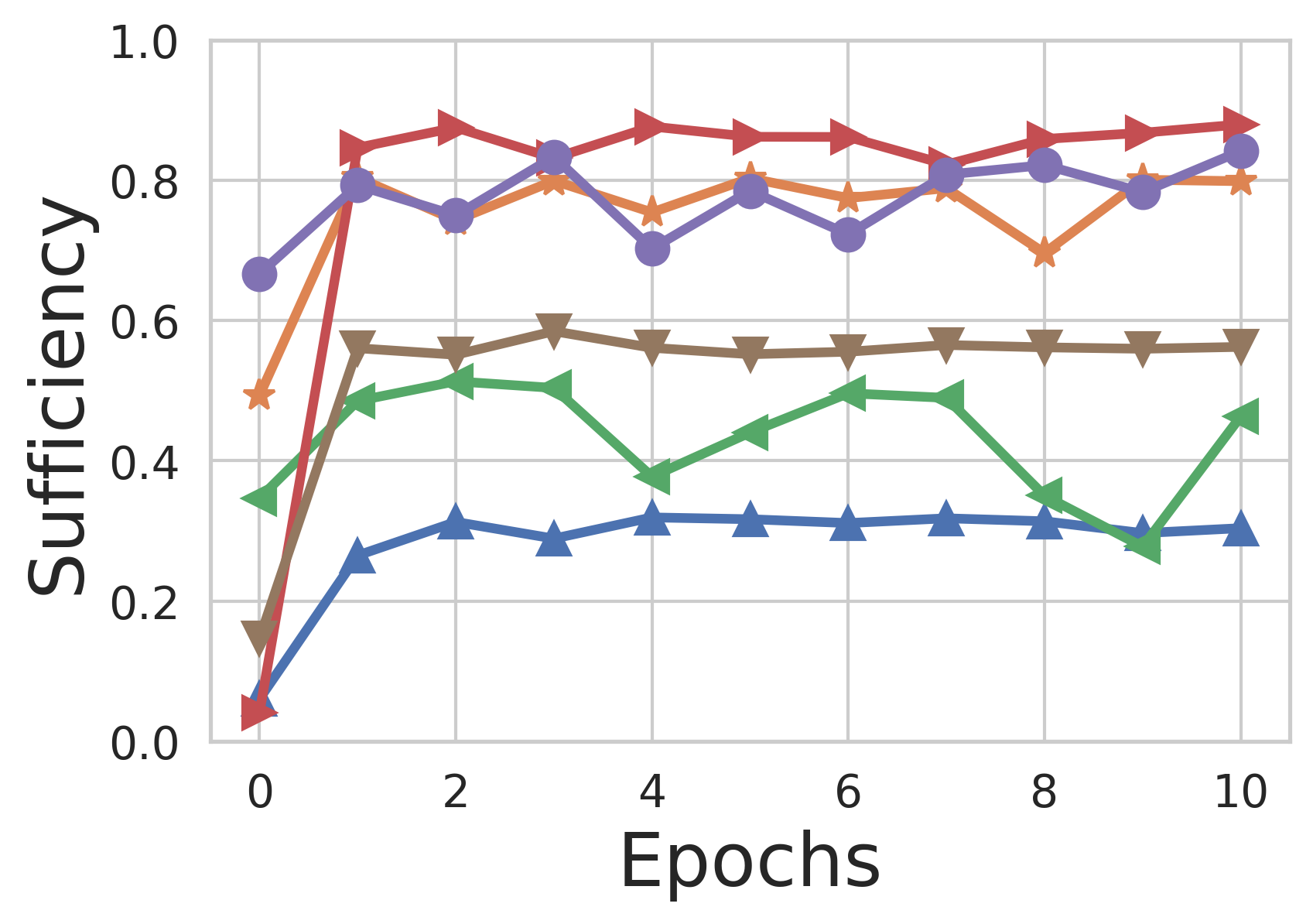} 

  \caption{Normalized sufficiency}
  \label{fig:lstm_epoch_sufficiency}
\end{subfigure}
\begin{subfigure}[]{0.38\textwidth}
  \centering
    \includegraphics[ width=\textwidth]{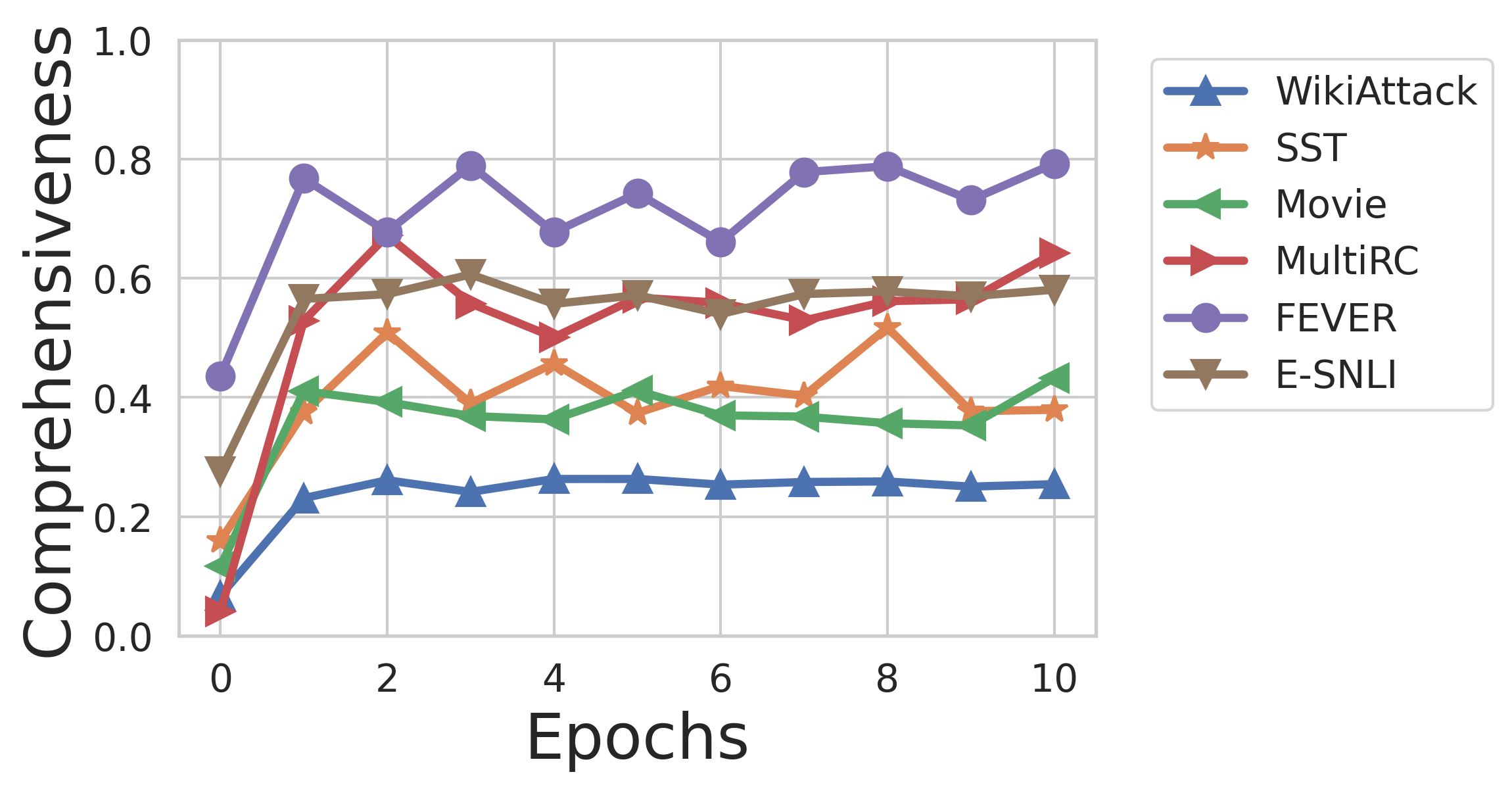} 

  \caption{Normalized comprehensiveness}
  \label{fig:lstm_epoch_comprehensiveness}
\end{subfigure}
\caption{Accuracy, sufficiency and comprehensiveness of LSTM models by training epoch.
}
\label{fig:lstm_epoch_wise_fidelity}
\end{figure*}

Finally, \figref{fig:lstm_epoch_wise_fidelity} shows the change in accuracy and fidelity over training epochs for the LSTM model. We again see that fidelity metrics have a tendency to fluctuate when accuracy has seemingly stabilized, such as \fever.

\begin{figure*}[h]
\centering
\begin{subfigure}[t]{0.30\textwidth}
  \centering
    \includegraphics[width=\linewidth]{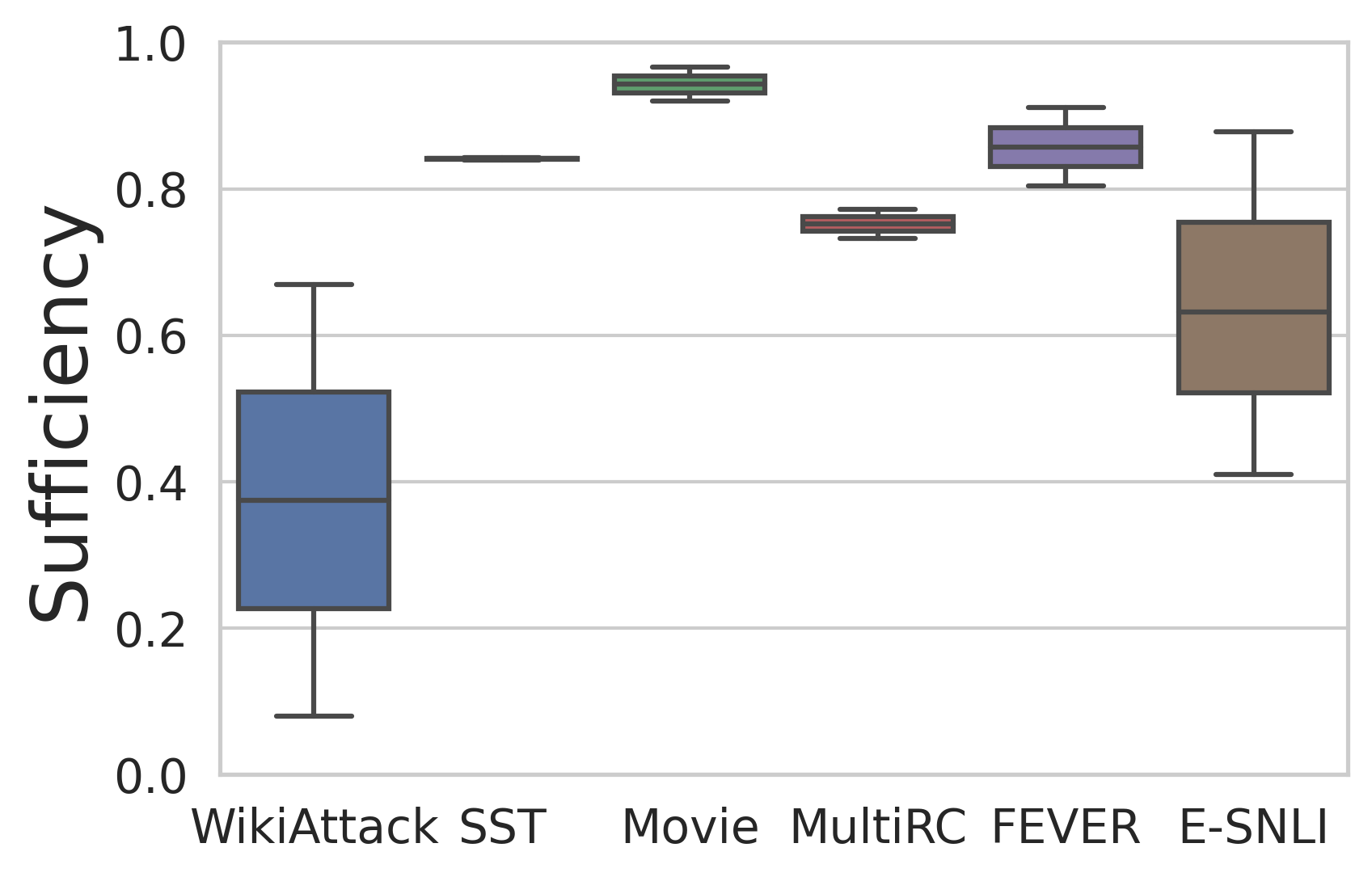}

  \caption{Normalized sufficiency}
  \label{fig:sufficiency_boxplot}
\end{subfigure}
\begin{subfigure}[t]{0.32\textwidth}
  \centering
    \includegraphics[width=\linewidth]{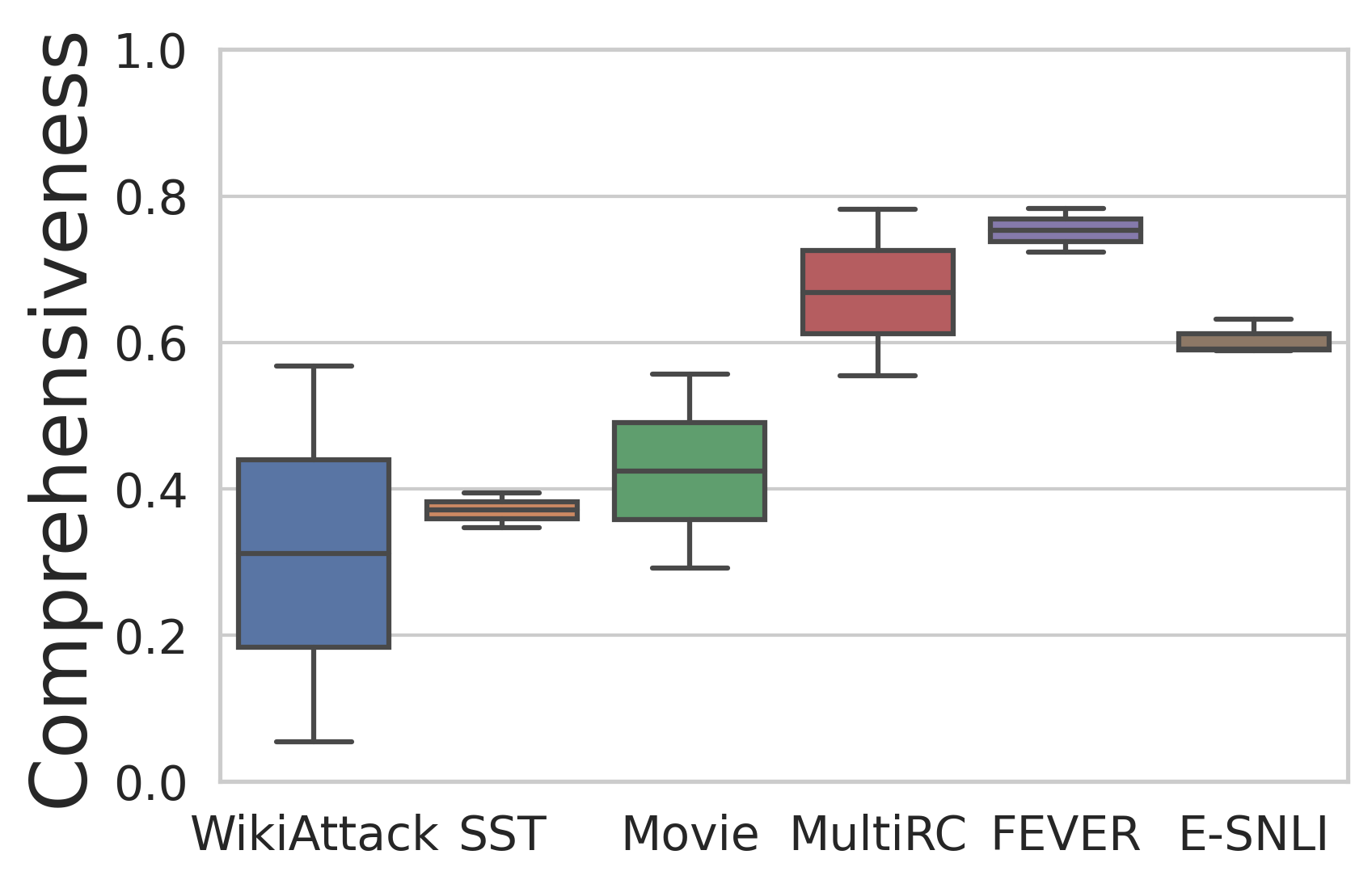}

  \caption{Normalized comprehensiveness}
  \label{fig:comprehensiveness_boxplot}
\end{subfigure}
\caption{Box plots of normalized fidelity metrics.
}
\label{fig:fidelity_boxplots}
\end{figure*}

\afterpage{\clearpage}

\section{Distribution of Fidelity Scores}

\figref{fig:fidelity_boxplots} shows box plots of normalized fidelity scores for the six datasets. We see a wide range of variances. \wikiattack and \esnli, the two datasets with assymmetric rationales, display the highest variance in sufficiency, while \wikiattack, \movie, and \multirc who relatively high variance in their comprehensiveness scores.

\end{document}